\documentclass[10pt,twocolumn,letterpaper]{article}

\usepackage[pagenumbers]{style/cvpr}             %

\usepackage[dvipsnames]{xcolor}
\usepackage[export]{adjustbox}

\definecolor{cvprblue}{rgb}{0.21,0.49,0.74}
\usepackage[pagebackref,breaklinks,colorlinks,citecolor=cvprblue]{hyperref}

\usepackage{subcaption}  %
\usepackage{makecell}   %
\usepackage{amsmath}
\usepackage{nicefrac}
\usepackage{tikz}
\usepackage{multirow}

\usepackage{xargs}      %
\usepackage{soul}       %
\usepackage{color}      %

\definecolor{LIGHTPINK}{RGB}{237,157,202}
\definecolor{LIGHTRED}{RGB}{210,121,121}
\definecolor{LIGHTORANGE}{RGB}{230,170,50}
\definecolor{LIGHTGOLD}{RGB}{210,194,121}
\definecolor{LIGHTGREEN}{RGB}{121,210,121}
\definecolor{LIGHTAQUA}{RGB}{121,206,210}
\definecolor{LIGHTBLUE}{RGB}{121,124,210}
\definecolor{LIGHTPURPLE}{RGB}{153,102,255}
\definecolor{RED}{RGB}{178,34,34}
\definecolor{GRAY}{RGB}{166,166,166}
\definecolor{WHITE}{RGB}{255,255,255}

\newcommand{\showifneeded}[1]{}  %

\newcommand{\showrevise}[1]{}  %

\newcommand{\todo}[1]{\showifneeded{
    \addcontentsline{toc}{subsection}{  %
        \protect\numberline{}           %
        \textcolor{RED}{[TODO] #1}}     %
    \textcolor{RED}{[TODO] \emph{#1}}}}  %

\newcommand{\revise}[1]{\showrevise{\textcolor{blue}{#1}}}

\newcommandx{\pet}[2][1=] 
    {\showifneeded{\setulcolor{LIGHTGREEN}{\ul{#1}} \textcolor{LIGHTGREEN}
    {[\textbf{Peter:} #2]}}}
\newcommandx{\guest}[3][1=]
    {\showifneeded{\setulcolor{LIGHTORANGE}{\ul{#1}} \textcolor{LIGHTORANGE} 
    {[\textbf{#2:} #3]}}}
\newcommandx{\yhg}[2][1=] 
    {\showifneeded{\setulcolor{LIGHTRED}{\ul{#1}} \textcolor{LIGHTRED}
    {[\textbf{yhg:} #2]}}}
\newcommandx{\jp}[2][1=] 
    {\showifneeded{\color{blue}[JP:#1]}}

\newcommand{\badge}[2]{\colorbox{#1}{\small\textcolor{WHITE}{\texttt{\textbf{#2}}}}}
\newcommand{\headerBadge}[2]{\hspace*{\fill}\badge{#1}{#2}}

\newcommand{\readyForFeedback}{\showifneeded{\headerBadge{LIGHTORANGE}{feedback requested}}}

\newcommand{\incomplete}{\showifneeded{\headerBadge{RED}{incomplete}}}

\newcommand{\figTeaser}[1]{
    \centering
    \vspace{-48pt}
    \vbox{%
    	\hsize\textwidth
    	\linewidth\hsize
    	\centering
            \large{$^1$Technical University of Munich\,\quad $^2$Adobe Research} \\ 
            \vspace{4pt}
    	\normalsize
    	\tt\href{https://peter-kocsis.github.io/LightIt/}{peter-kocsis.github.io/LightIt/}
    }
    \vspace{4pt}
    \includegraphics[width=#1\textwidth]{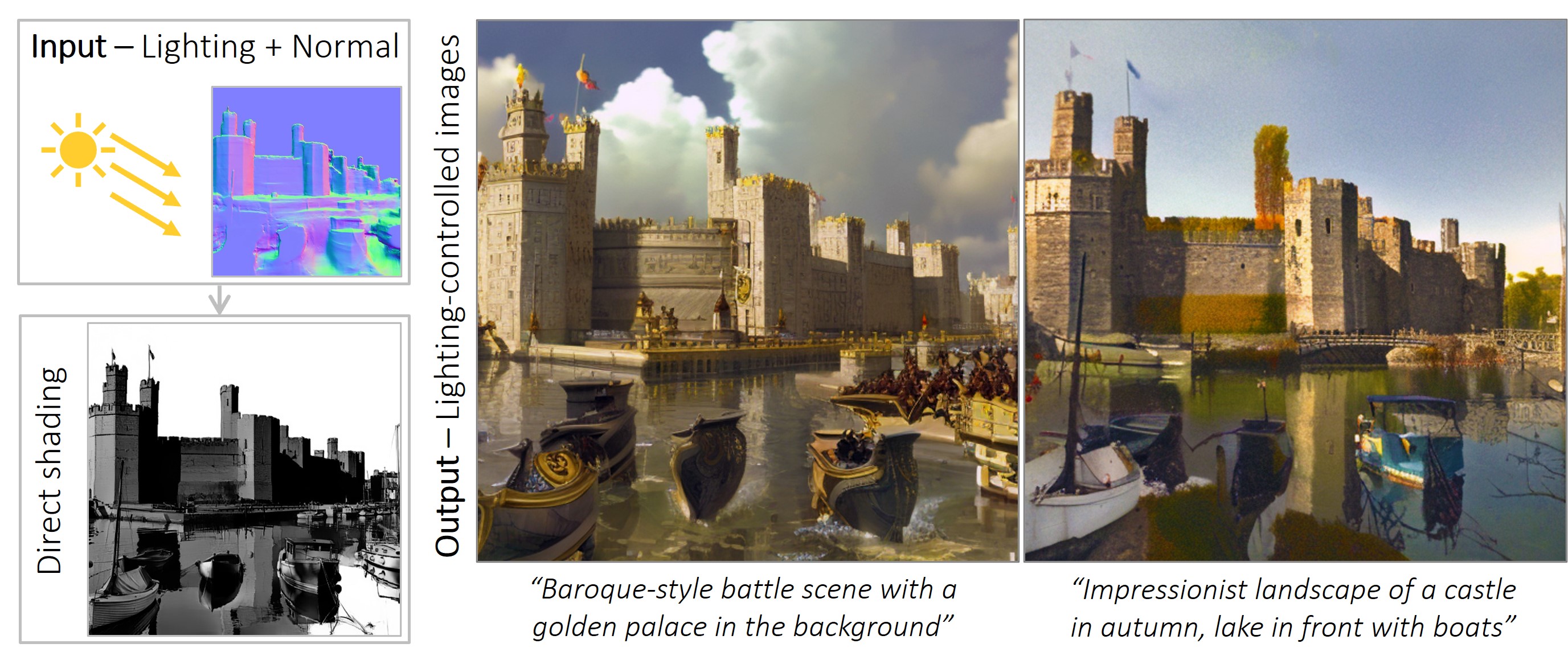}

    \vspace{-9pt}
    \captionsetup{type=figure}\caption{
    \textbf{Teaser.} 
    We present \emph{LightIt}, a method for explicit lighting control of text-guided image generation. 
    Given a normal map of the desired geometry and light direction with a solid angle, we introduce a method to predict direct shading, which is then used to generate high-quality images with coherent lighting. 
    \readyForFeedback}
    \label{fig:teaser}
    \vspace{3pt}
}
\makeatletter
\apptocmd\@maketitle{{\figTeaser{}\par}}{}{}
\makeatother

\newcommand{\figShadingEstimation}{
\begin{figure*}[t]
    \centering
    \includegraphics[width=\textwidth]{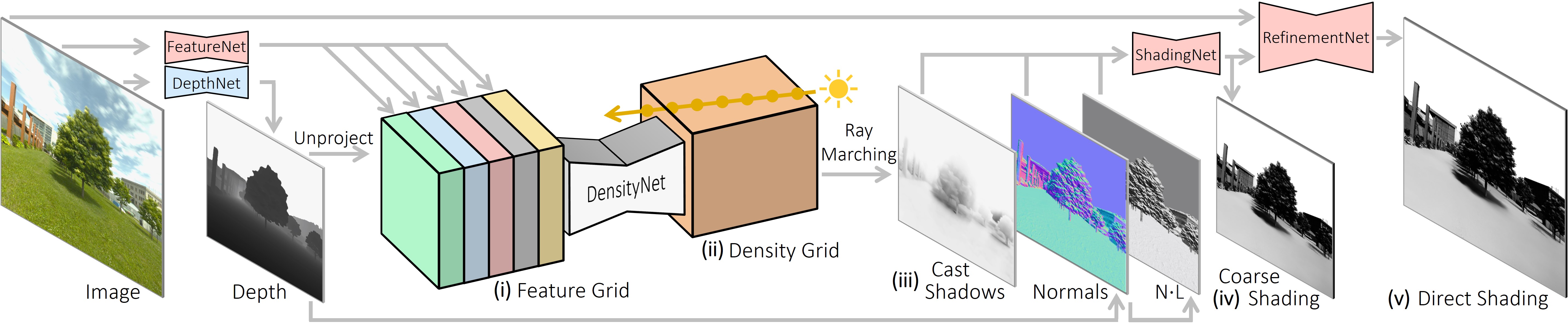}

    \vspace{-6pt}
    \caption{\textbf{Shading Estimation}. 
    We estimate the direct shading of a single image. 
    (i) We predict image features (FeatureNet) and unproject them to a 3D feature grid in NDC space. 
    (ii) We predict a density field from the features (DensityNet). 
    (iii) Given the sun's direction and solid angle, we trace rays toward the lightsource to obtain a coarse shadow map. 
    (iv) Using the shadows and N-dot-L shading information, we predict a coarse shading map (ShadingNet). 
    (v) We refine the shading map to get our direct shading (RefinementNet). 
    \readyForFeedback}
    \label{fig:method:shading_estimation}
    \vspace{-15pt}
\end{figure*}
}

\newcommand{\figDatasetPipeline}{
\begin{figure}[t]
    \centering
    \includegraphics[width=\columnwidth]{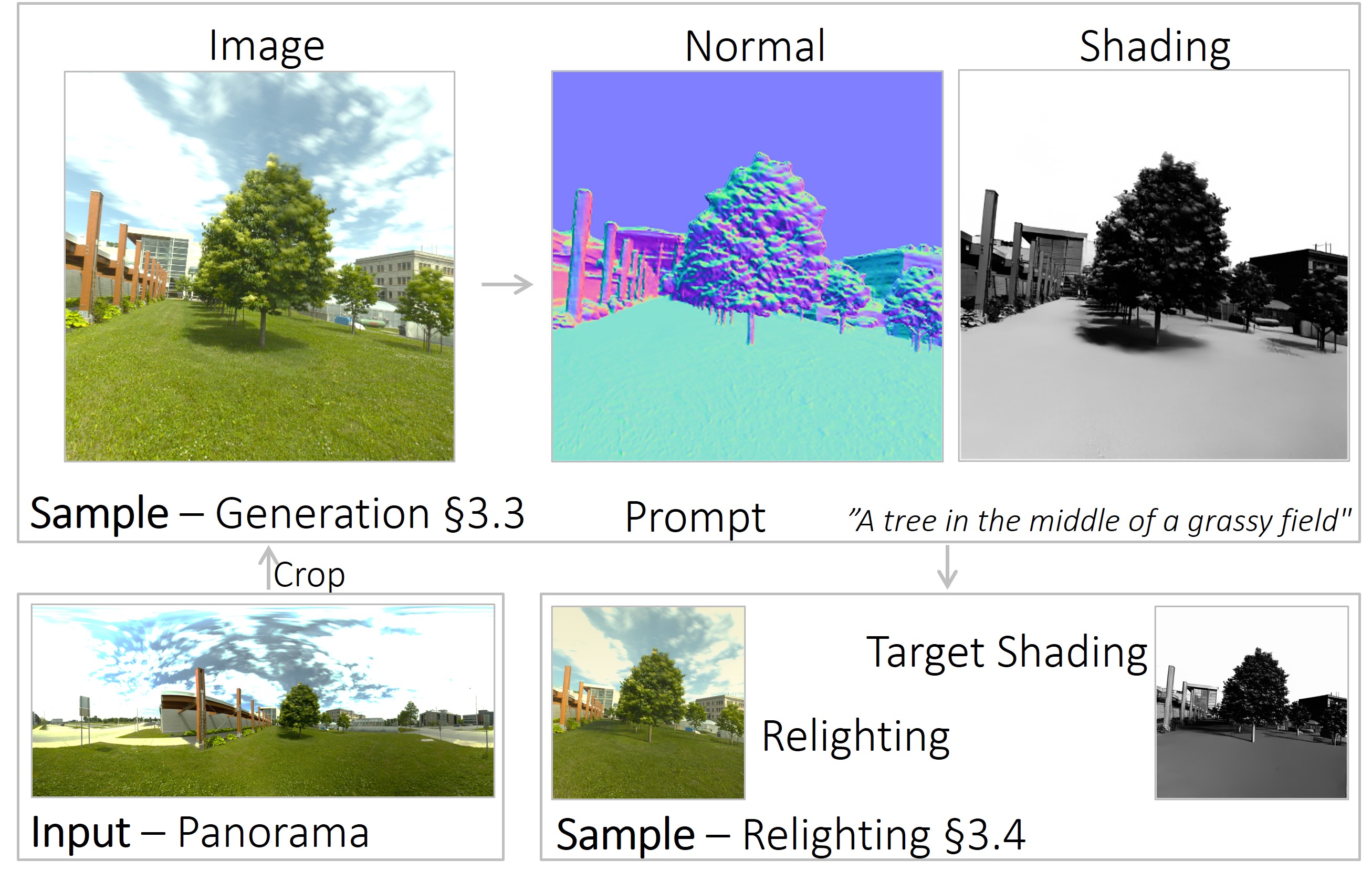}

    \vspace{-6pt}
    \caption{\textbf{Dataset Generation Pipeline}. 
    We generate a dataset using the Outdoor Laval dataset \cite{hold2019deep}. 
    We randomly crop images from the panoramas and automatically predict normal, shading, and caption (\cref{sec:method:dataset}).
    For our relighting experiments (\cref{sec:method:relighting}), we extend the dataset with relit images using OutCast \cite{griffiths2022outcast}. 
    \readyForFeedback}
    \label{fig:method:dataset_pipeline}
    \vspace{-9pt}
\end{figure}
}

\newcommand{\figTrainingPipeline}{
\begin{figure}[t]
    \centering
    \includegraphics[width=\columnwidth]{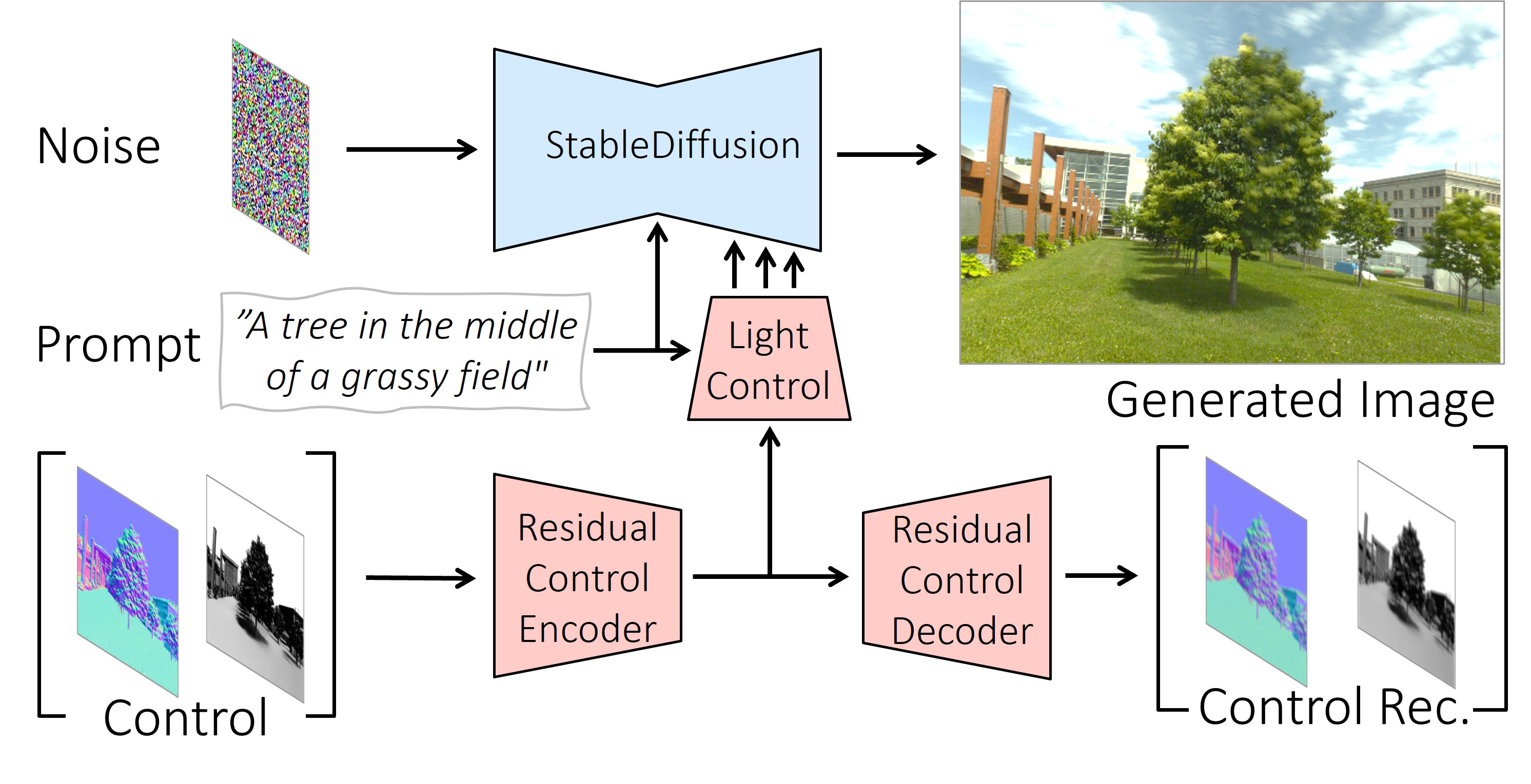}

    \vspace{-9pt}
    \caption{\textbf{Model Overview}. 
    To generate lighting-controlled images, we train a light control module similar to \cite{zhang2023controlnet}, conditioned on normal and shading estimation. 
    We use a custom Residual Control Encoder to encode the control signal for the ControlNet. 
    Adding a Residual Control Decoder with a reconstruction loss ensures the full control signal is present in the encoded signal.
    \readyForFeedback}
    \label{fig:method:training_pipeline}
    \vspace{-15pt}
\end{figure}
}

\newcommand{\figConsistentImageSynthesis}{
\begin{figure*}[t]
    \centering
    \setlength\tabcolsep{1.25pt}
    \resizebox{\textwidth}{!}{
    \begin{tabular}{c|ccc}
        \begin{tabular}{c}
            \rotatebox{90}{Shading}
            \includegraphics[width=0.1175\textwidth]{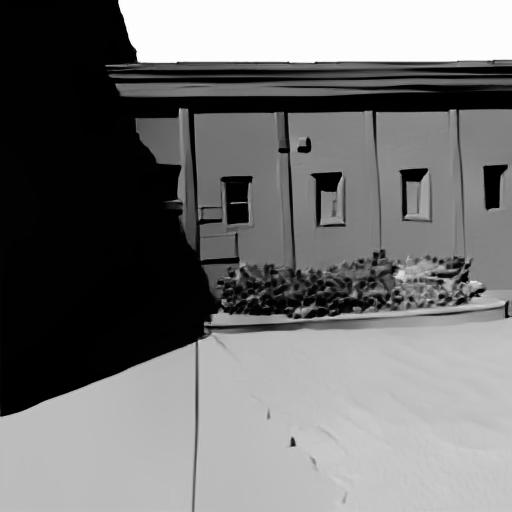}\\
            \rotatebox{90}{Normal}
            \includegraphics[width=0.1175\textwidth]{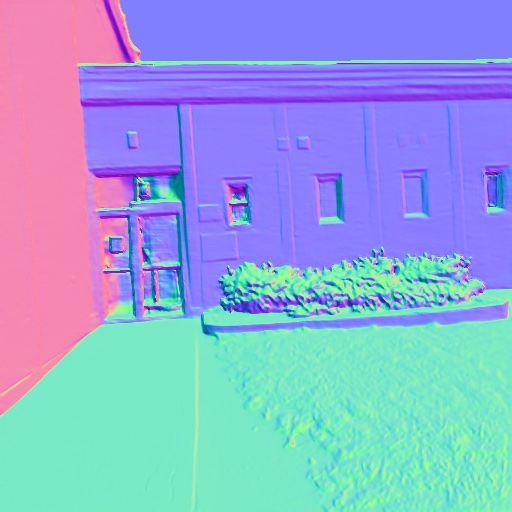}\\
        \end{tabular}
        & 
        \begin{tabular}{c}
            \includegraphics[width=0.24\textwidth]{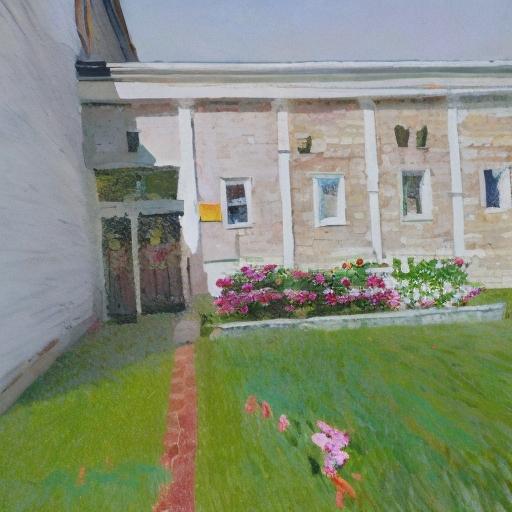}
        \end{tabular}
        & 
        \begin{tabular}{c}
            \includegraphics[width=0.24\textwidth]{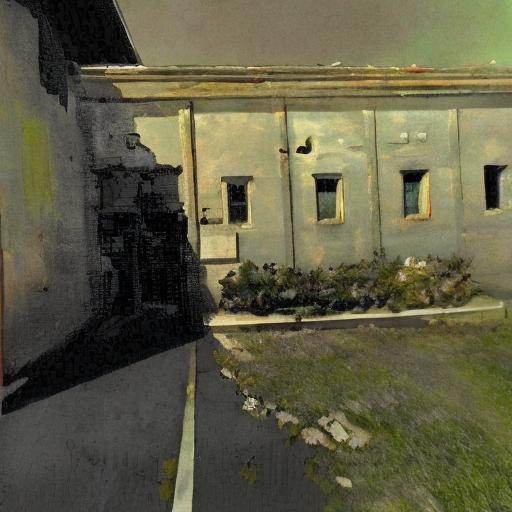}
        \end{tabular}
        & 
        \begin{tabular}{c}
            \includegraphics[width=0.24\textwidth]{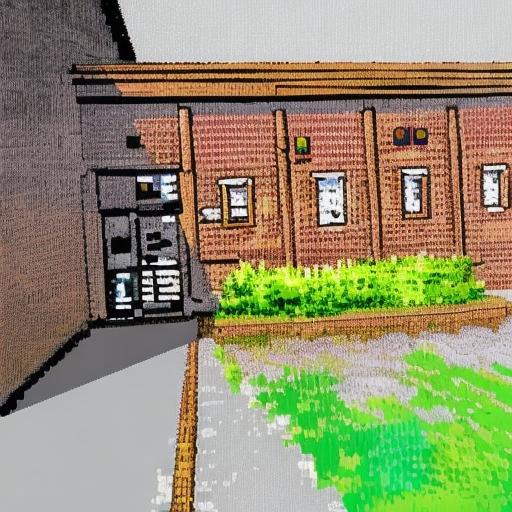}
        \end{tabular}\\
        Conditioning 
        & 
        \makecell{
        \textit{Impressionist painting of} \\ 
        \textit{an english row house}} 
        & 
        \makecell{
        \textit{Gothic painting of an } \\
        \textit{abandoned house in full moon}}
        &
        \makecell{
        \textit{Pixel art of an 8-bit video} \\
        \textit{game medieval house}}
    \end{tabular}}

    \vspace{-6pt}
    \caption{\textbf{Image Synthesis with Consistent Lighting}. 
    Our generated images feature consistent lighting aligned with the target shading for diverse text prompts. 
    \readyForFeedback}
    \label{fig:exp:consistent_image_synthesis}
    \vspace{-12pt}
\end{figure*}
}

\newcommand{\figIDControllableImageSynthesis}{
\begin{figure*}[t]
    \centering
    \setlength\tabcolsep{1.25pt}
    \resizebox{\textwidth}{!}{
    \fboxsep=0pt
    \begin{tabular}{c|c|ccc}
        \fbox{\includegraphics[width=0.19\textwidth]{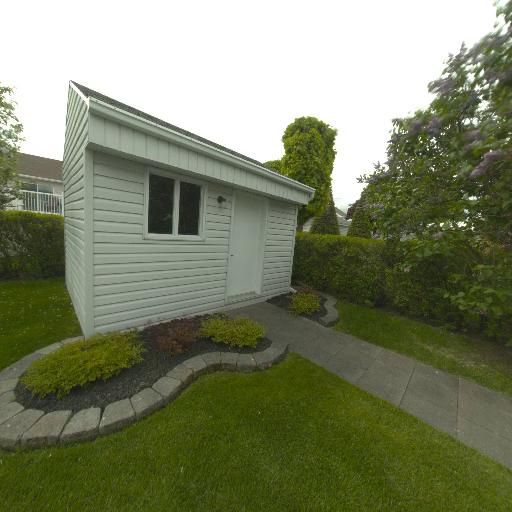}}
        & 
        \begin{tabular}[b]{cc}
            \makecell{
                \footnotesize{\textit{A small white shed}} \\
                \footnotesize{\textit{sitting on a green field}}} \\
            \fbox{\includegraphics[width=0.14\textwidth]{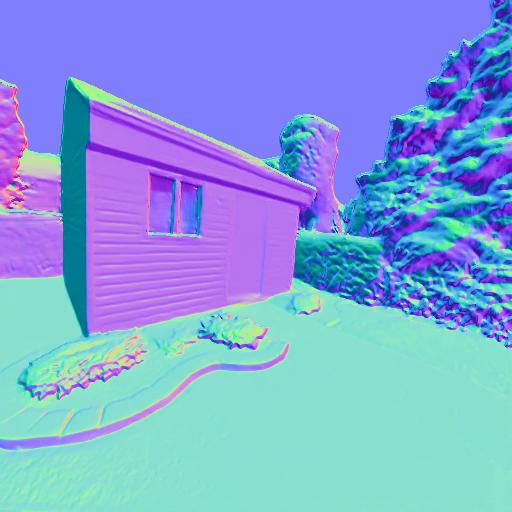}} 
        \end{tabular}
        &
        \begin{tikzpicture}[every node/.style={anchor=north east,inner sep=0pt},x=-1pt, y=-1pt,]  
             \node (fig1) at (0,0)
               {\fbox{\includegraphics[width=0.19\textwidth]{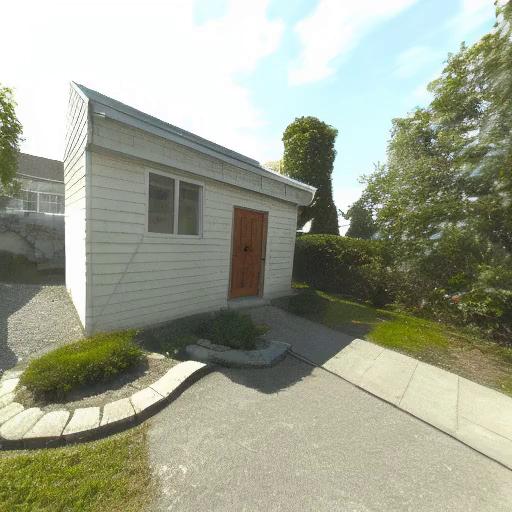}}};   
             \node (fig2) at (-6,-6)
               {\fbox{\includegraphics[width=0.07\textwidth]{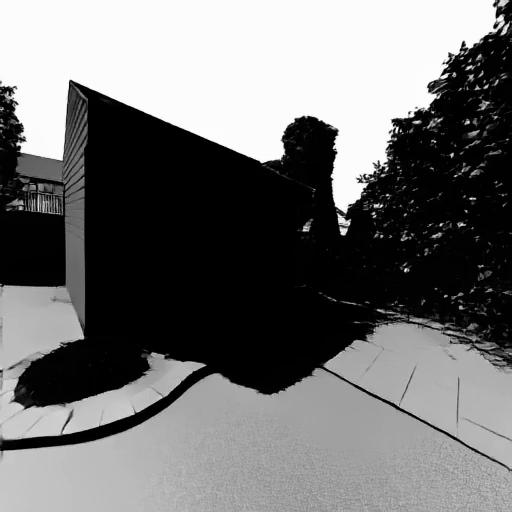}}};
        \end{tikzpicture}
        &
        \begin{tikzpicture}[every node/.style={anchor=north east,inner sep=0pt},x=-1pt, y=-1pt,]  
             \node (fig1) at (0,0)
               {\fbox{\includegraphics[width=0.19\textwidth]{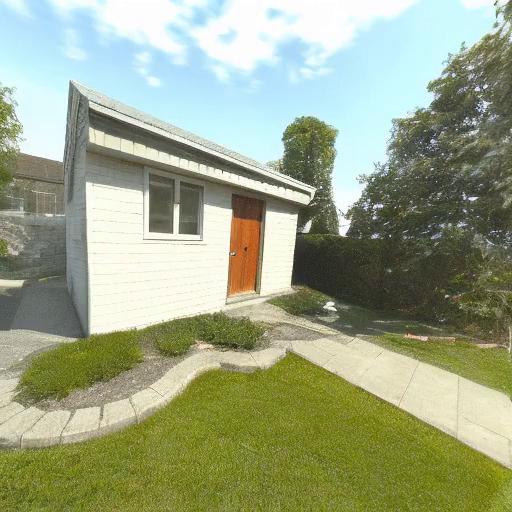}}};   
             \node (fig2) at (-6,-6)
               {\fbox{\includegraphics[width=0.07\textwidth]{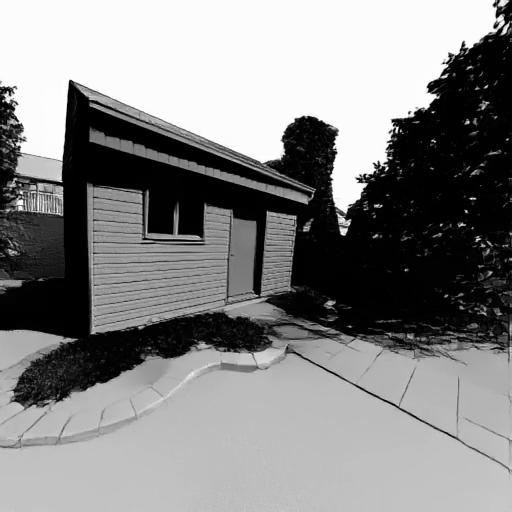}}};
        \end{tikzpicture}
        &
        \begin{tikzpicture}[every node/.style={anchor=north east,inner sep=0pt},x=-1pt, y=-1pt,]  
             \node (fig1) at (0,0)
               {\fbox{\includegraphics[width=0.19\textwidth]{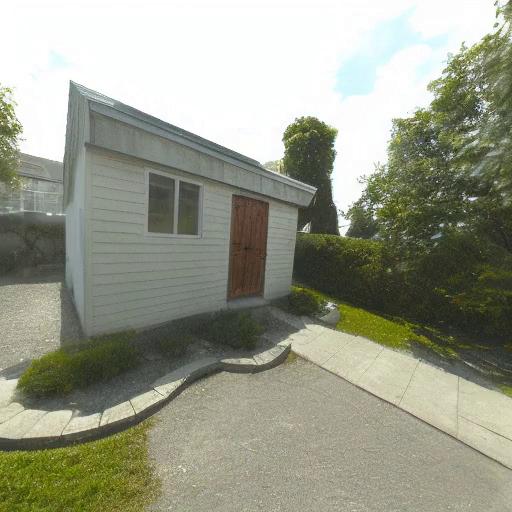}}};   
             \node (fig2) at (-6,-6)
               {\fbox{\includegraphics[width=0.07\textwidth]{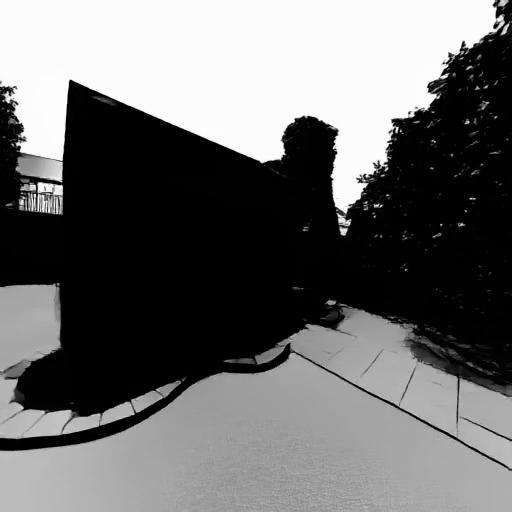}}};
        \end{tikzpicture}
        \\

        \fbox{\includegraphics[width=0.19\textwidth]{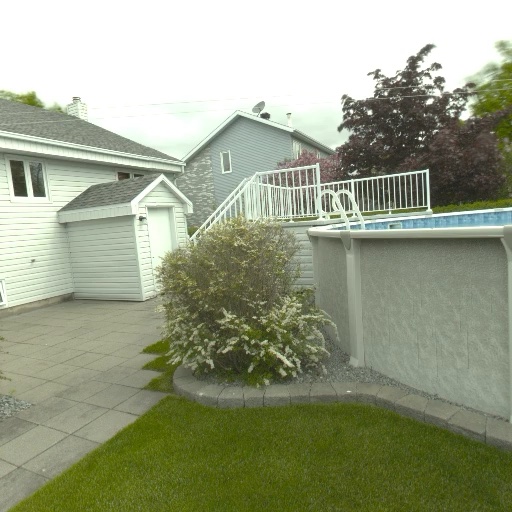}}
        & 
        \begin{tabular}[b]{cc}
            \makecell{
                \footnotesize{\textit{A backyard with a}} \\
                \footnotesize{\textit{pool and a fence}}} \\
            \fbox{\includegraphics[width=0.14\textwidth]{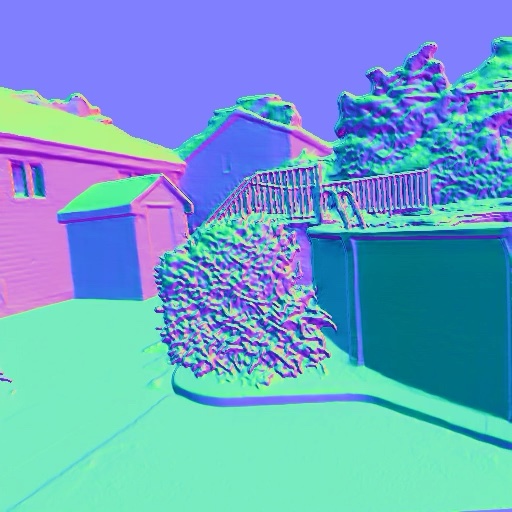}} 
        \end{tabular}
        &
        \begin{tikzpicture}[every node/.style={anchor=north east,inner sep=0pt},x=-1pt, y=-1pt,]  
             \node (fig1) at (0,0)
               {\fbox{\includegraphics[width=0.19\textwidth]{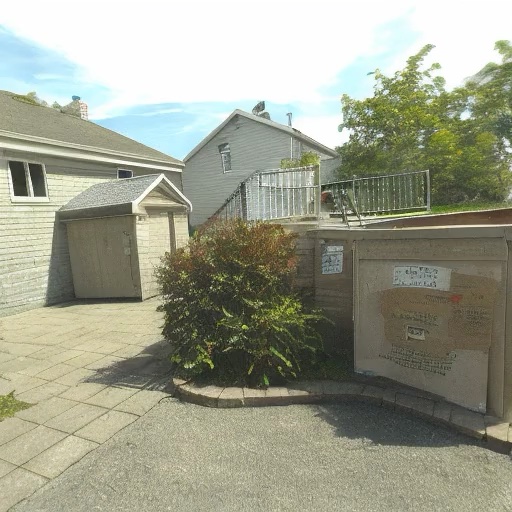}}};   
             \node (fig2) at (-6,-6)
               {\fbox{\includegraphics[width=0.07\textwidth]{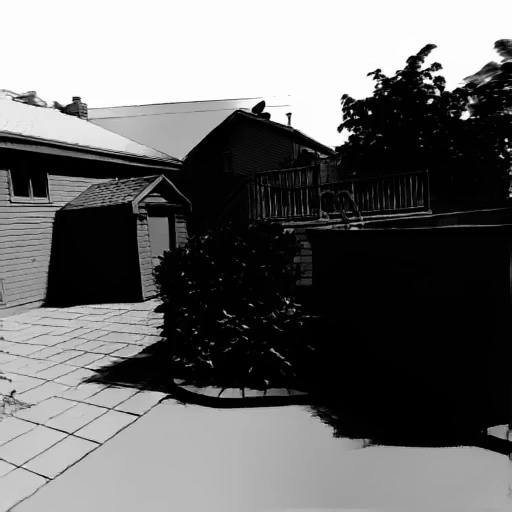}}};
        \end{tikzpicture}
        &
        \begin{tikzpicture}[every node/.style={anchor=north east,inner sep=0pt},x=-1pt, y=-1pt,]  
             \node (fig1) at (0,0)
               {\fbox{\includegraphics[width=0.19\textwidth]{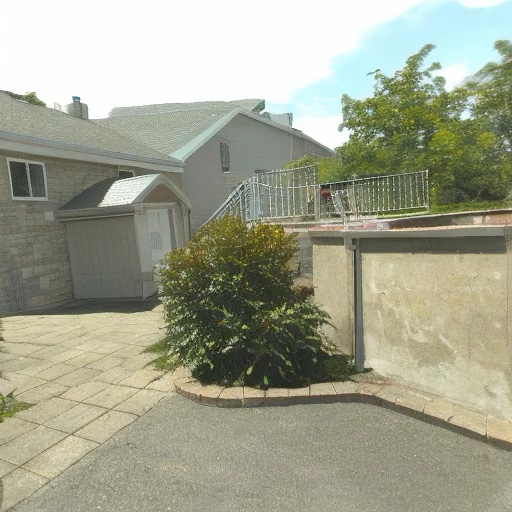}}};   
             \node (fig2) at (-6,-6)
               {\fbox{\includegraphics[width=0.07\textwidth]{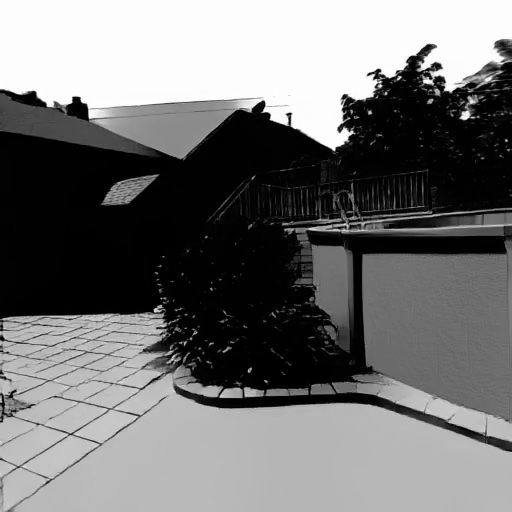}}};
        \end{tikzpicture}
        &
        \begin{tikzpicture}[every node/.style={anchor=north east,inner sep=0pt},x=-1pt, y=-1pt,]  
             \node (fig1) at (0,0)
               {\fbox{\includegraphics[width=0.19\textwidth]{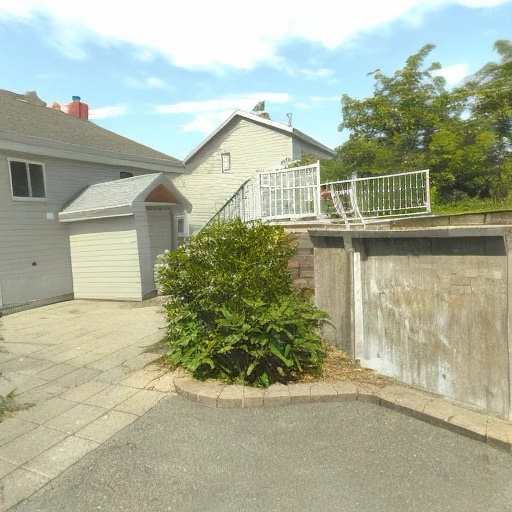}}};   
             \node (fig2) at (-6,-6)
               {\fbox{\includegraphics[width=0.07\textwidth]{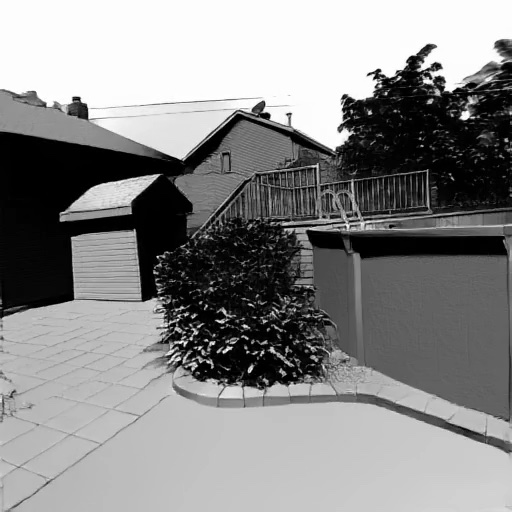}}};
        \end{tikzpicture}
        \\
        
        Original Image & Input & Lighting 1 & Lighting 2 & Lighting 3
    \end{tabular}}

    \vspace{-3pt}
    \caption{\textbf{In-Domain Image Synthesis with Controllable Lighting}. 
    We can synthesize images under various lighting conditions. 
    \readyForFeedback}
    \label{fig:exp:id_controllable_image_synthesis}
    \vspace{-6pt}
\end{figure*}
}

\newcommand{\figODControllableImageSynthesis}{
\begin{figure*}[t]
    \centering
    \resizebox{0.97\textwidth}{!}{
    \setlength\tabcolsep{1.25pt}
    \fboxsep=0pt
    \begin{tabular}{ccccc}
        \rotatebox{90}{\textit{Impressionist painting}} &
        \begin{tikzpicture}[every node/.style={anchor=north west,inner sep=0pt},x=1pt, y=-1pt,]  
             \node (fig1) at (0,0)
               {\fbox{\includegraphics[width=0.23\textwidth]{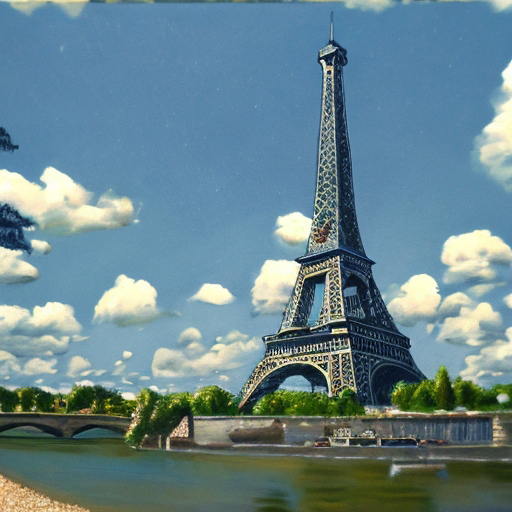}}};   
             \node (fig2) at (-6,-6)
               {\fbox{\includegraphics[width=0.08\textwidth]{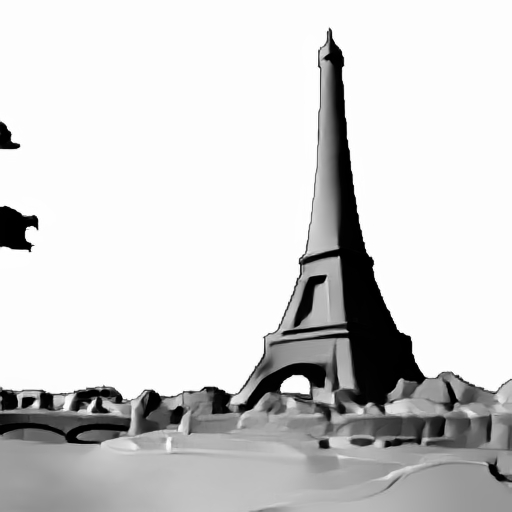}}};
        \end{tikzpicture} &
        \begin{tikzpicture}[every node/.style={anchor=north west,inner sep=0pt},x=1pt, y=-1pt,]  
             \node (fig1) at (0,0)
               {\fbox{\includegraphics[width=0.23\textwidth]{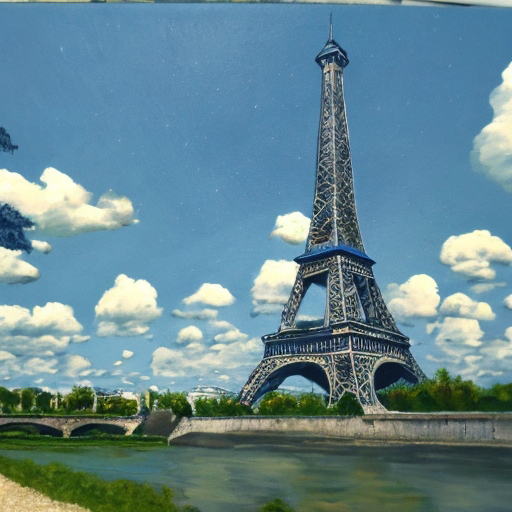}}};   
             \node (fig2) at (-6,-6)
               {\fbox{\includegraphics[width=0.08\textwidth]{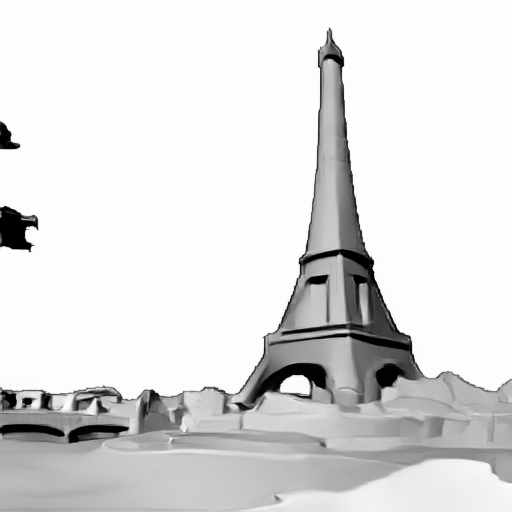}}};
        \end{tikzpicture} &
        \begin{tikzpicture}[every node/.style={anchor=north west,inner sep=0pt},x=1pt, y=-1pt,]  
             \node (fig1) at (0,0)
               {\fbox{\includegraphics[width=0.23\textwidth]{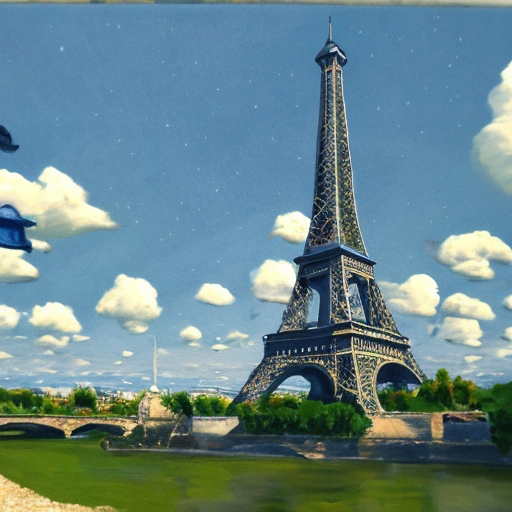}}};   
             \node (fig2) at (-6,-6)
               {\fbox{\includegraphics[width=0.08\textwidth]{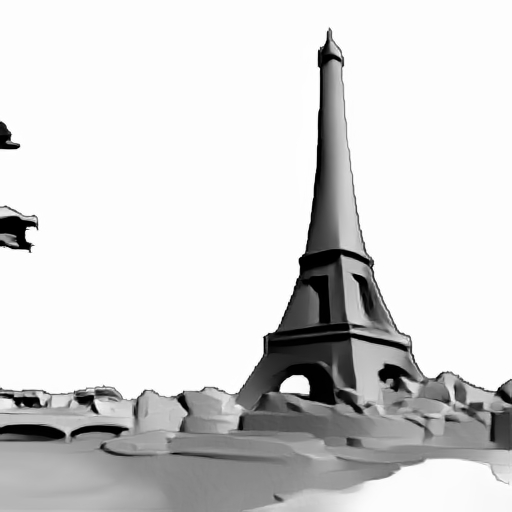}}};
        \end{tikzpicture} &
        \begin{tikzpicture}[every node/.style={anchor=north west,inner sep=0pt},x=1pt, y=-1pt,]  
             \node (fig1) at (0,0)
               {\fbox{\includegraphics[width=0.23\textwidth]{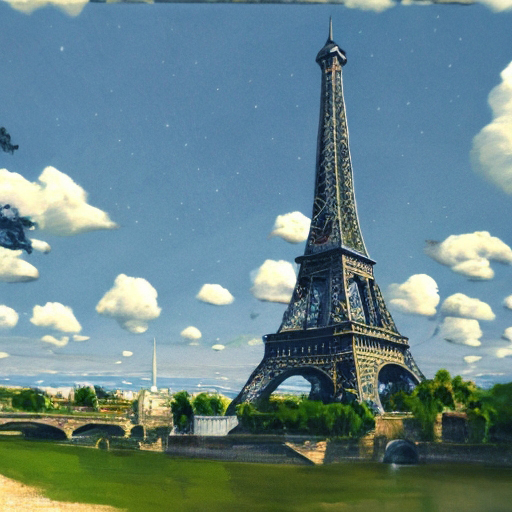}}};   
             \node (fig2) at (-6,-6)
               {\fbox{\includegraphics[width=0.08\textwidth]{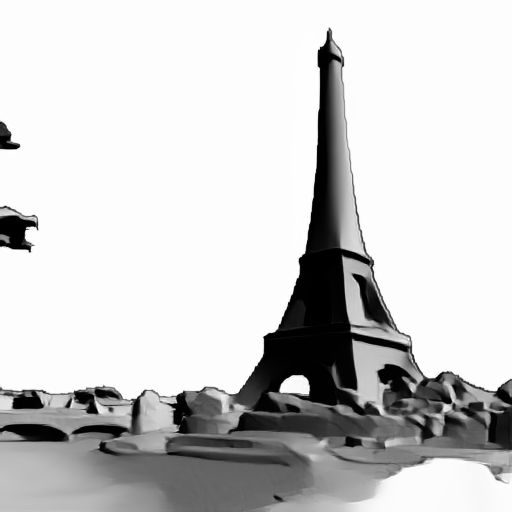}}};
        \end{tikzpicture}
        \\

        \rotatebox{90}{\textit{Medieval painting}} &
        \begin{tikzpicture}[every node/.style={anchor=north west,inner sep=0pt},x=1pt, y=-1pt,]  
             \node (fig1) at (0,0)
               {\fbox{\includegraphics[width=0.23\textwidth]{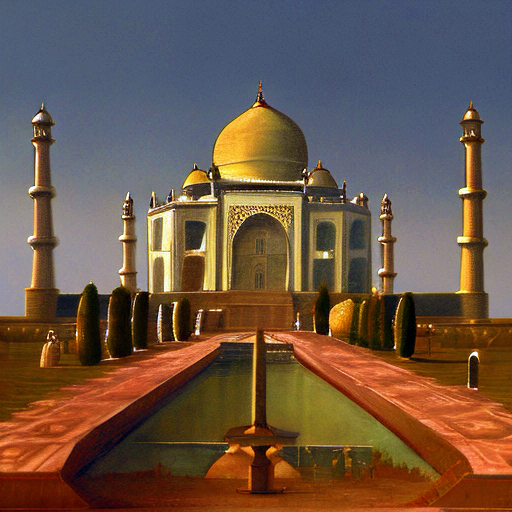}}};   
             \node (fig2) at (-6,-6)
               {\fbox{\includegraphics[width=0.08\textwidth]{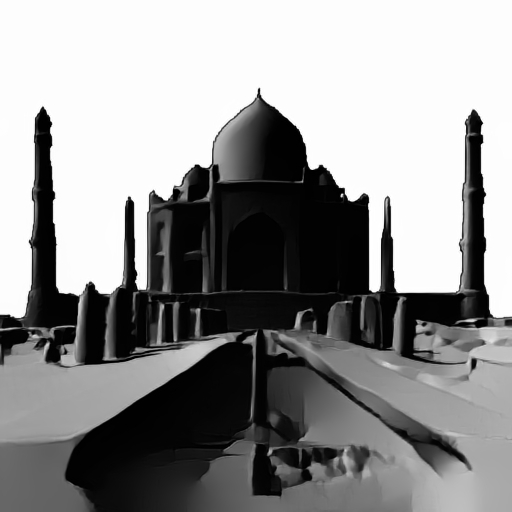}}};
        \end{tikzpicture} &
        \begin{tikzpicture}[every node/.style={anchor=north west,inner sep=0pt},x=1pt, y=-1pt,]  
             \node (fig1) at (0,0)
               {\fbox{\includegraphics[width=0.23\textwidth]{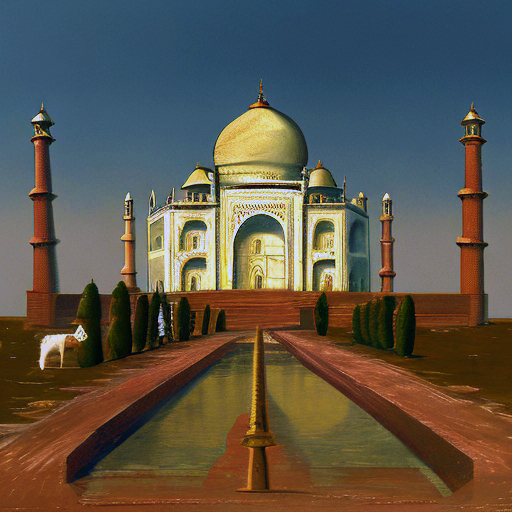}}};   
             \node (fig2) at (-6,-6)
               {\fbox{\includegraphics[width=0.08\textwidth]{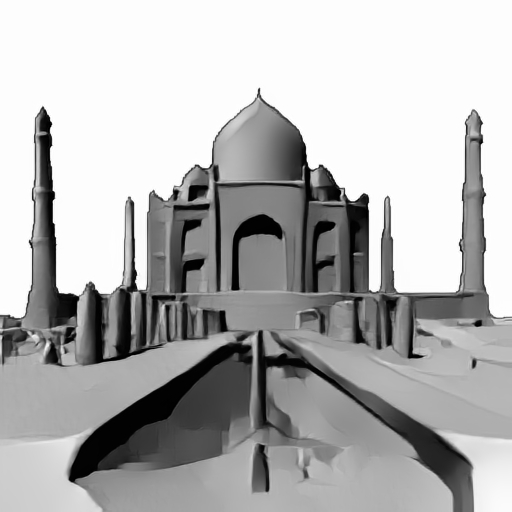}}};
        \end{tikzpicture} &
        \begin{tikzpicture}[every node/.style={anchor=north west,inner sep=0pt},x=1pt, y=-1pt,]  
             \node (fig1) at (0,0)
               {\fbox{\includegraphics[width=0.23\textwidth]{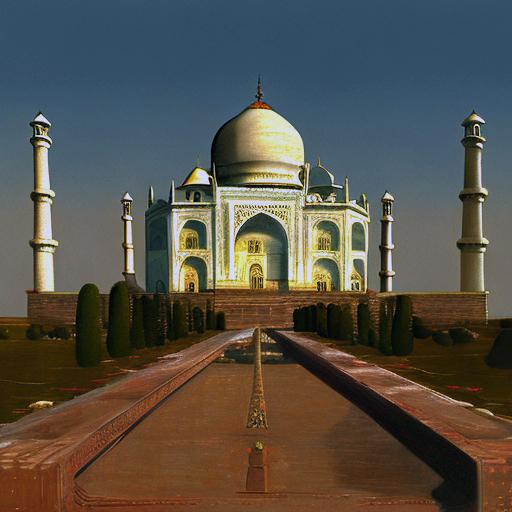}}};   
             \node (fig2) at (-6,-6)
               {\fbox{\includegraphics[width=0.08\textwidth]{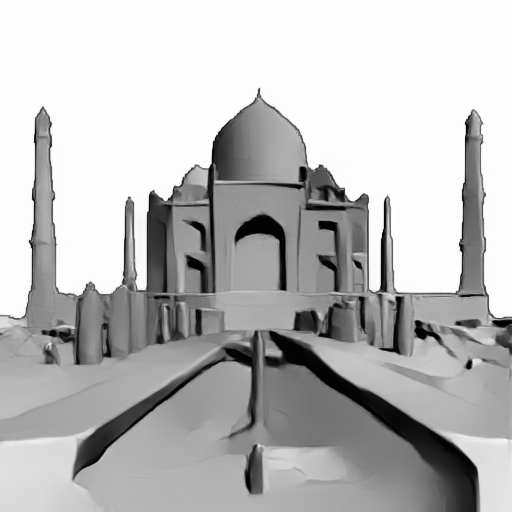}}};
        \end{tikzpicture} &
        \begin{tikzpicture}[every node/.style={anchor=north west,inner sep=0pt},x=1pt, y=-1pt,]  
             \node (fig1) at (0,0)
               {\fbox{\includegraphics[width=0.23\textwidth]{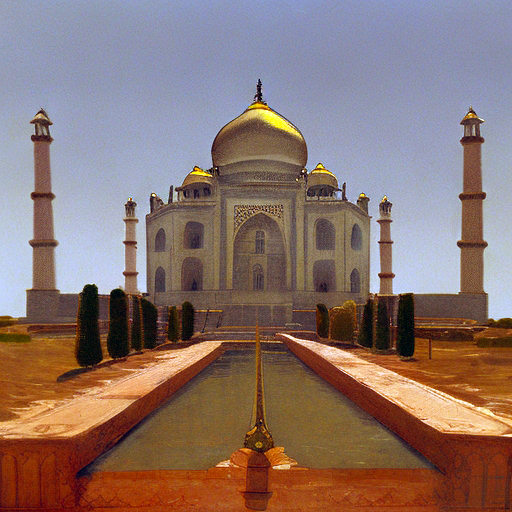}}};   
             \node (fig2) at (-6,-6)
               {\fbox{\includegraphics[width=0.08\textwidth]{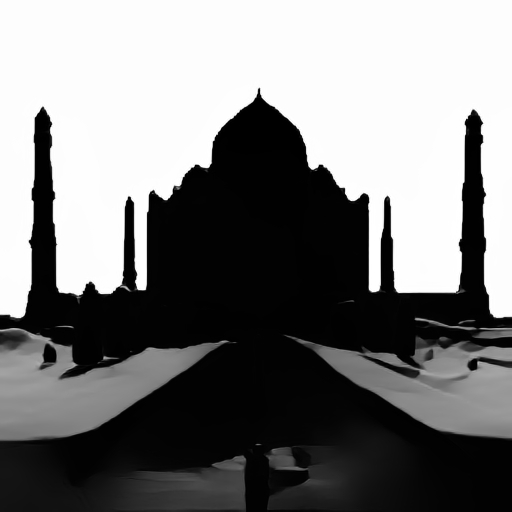}}};
        \end{tikzpicture}
        \\

        \rotatebox{90}{\textit{Drawing}} &
        \begin{tikzpicture}[every node/.style={anchor=north west,inner sep=0pt},x=1pt, y=-1pt,]  
             \node (fig1) at (0,0)
               {\fbox{\includegraphics[width=0.23\textwidth]{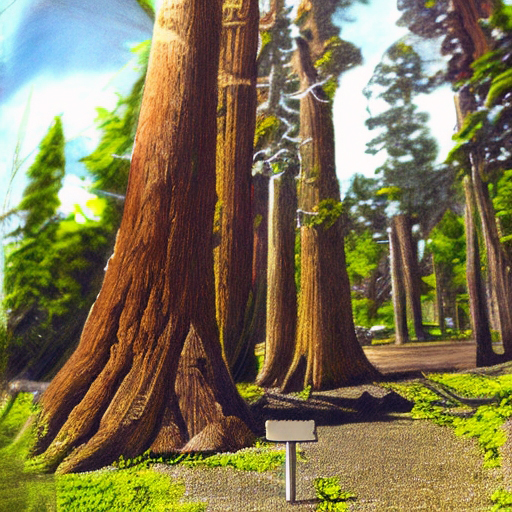}}};   
             \node (fig2) at (-6,-6)
               {\fbox{\includegraphics[width=0.08\textwidth]{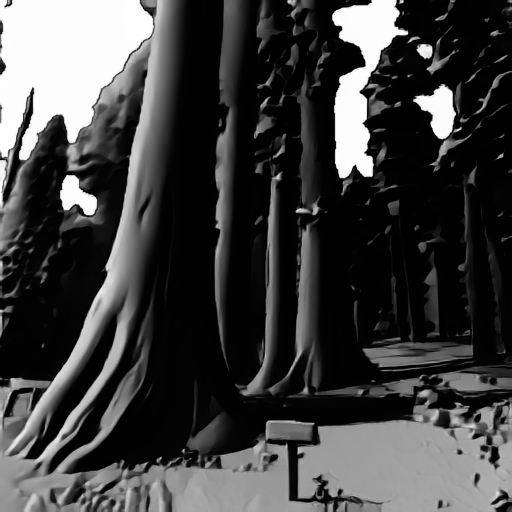}}};
        \end{tikzpicture} &
        \begin{tikzpicture}[every node/.style={anchor=north west,inner sep=0pt},x=1pt, y=-1pt,]  
             \node (fig1) at (0,0)
               {\fbox{\includegraphics[width=0.23\textwidth]{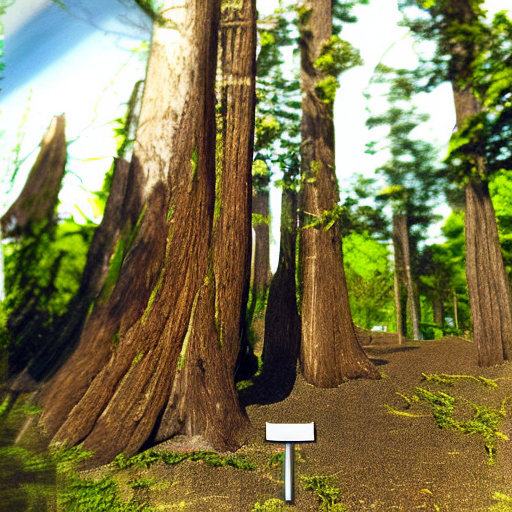}}};   
             \node (fig2) at (-6,-6)
               {\fbox{\includegraphics[width=0.08\textwidth]{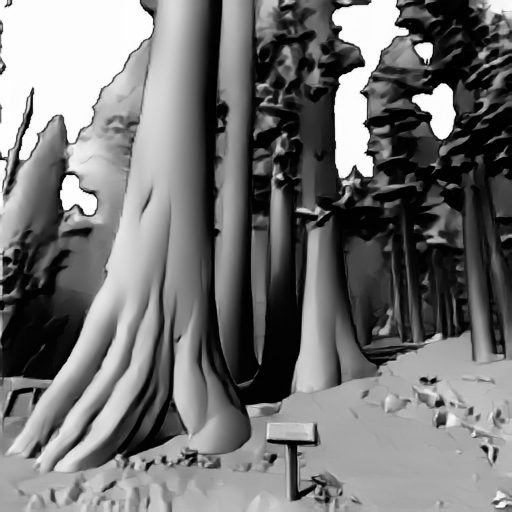}}};
        \end{tikzpicture} &
        \begin{tikzpicture}[every node/.style={anchor=north west,inner sep=0pt},x=1pt, y=-1pt,]  
             \node (fig1) at (0,0)
               {\fbox{\includegraphics[width=0.23\textwidth]{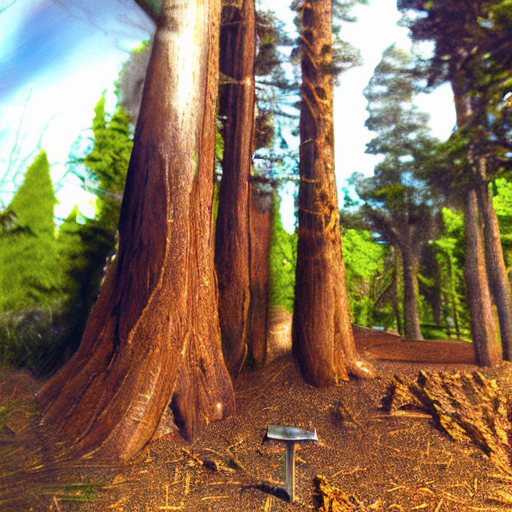}}};   
             \node (fig2) at (-6,-6)
               {\fbox{\includegraphics[width=0.08\textwidth]{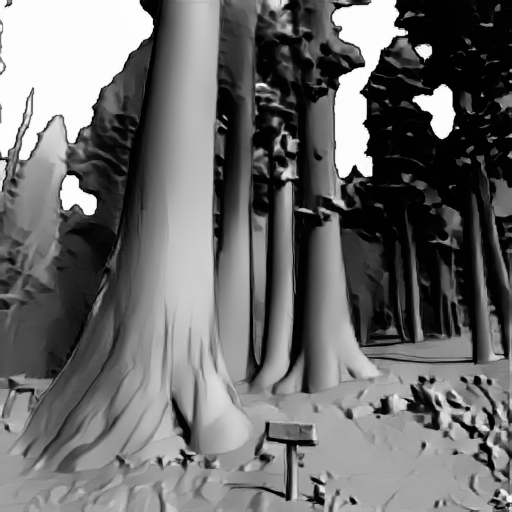}}};
        \end{tikzpicture} &
        \begin{tikzpicture}[every node/.style={anchor=north west,inner sep=0pt},x=1pt, y=-1pt,]  
             \node (fig1) at (0,0)
               {\fbox{\includegraphics[width=0.23\textwidth]{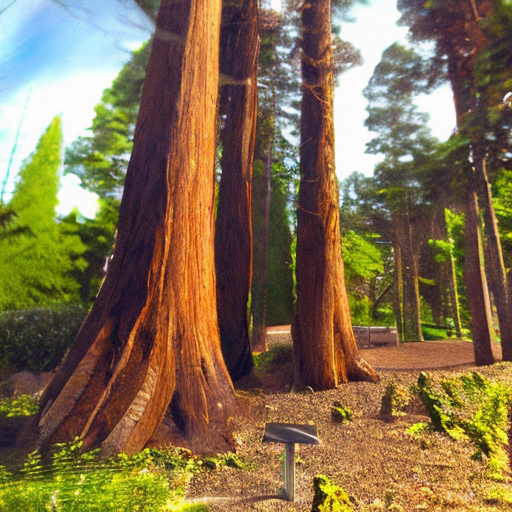}}};   
             \node (fig2) at (-6,-6)
               {\fbox{\includegraphics[width=0.08\textwidth]{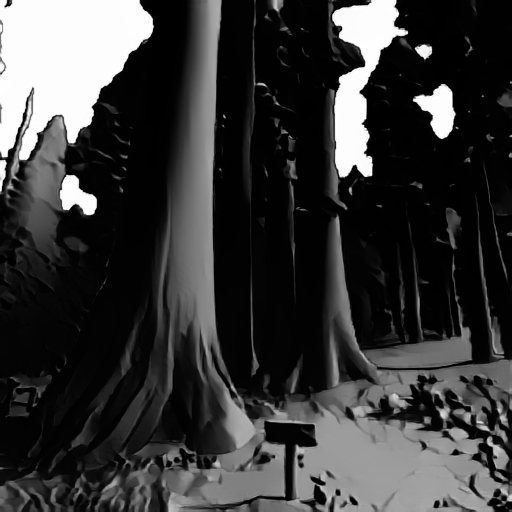}}};
        \end{tikzpicture}
        \\

        \rotatebox{90}{\textit{Image}} &
        \begin{tikzpicture}[every node/.style={anchor=north west,inner sep=0pt},x=1pt, y=-1pt,]  
             \node (fig1) at (0,0)
               {\fbox{\includegraphics[width=0.23\textwidth]{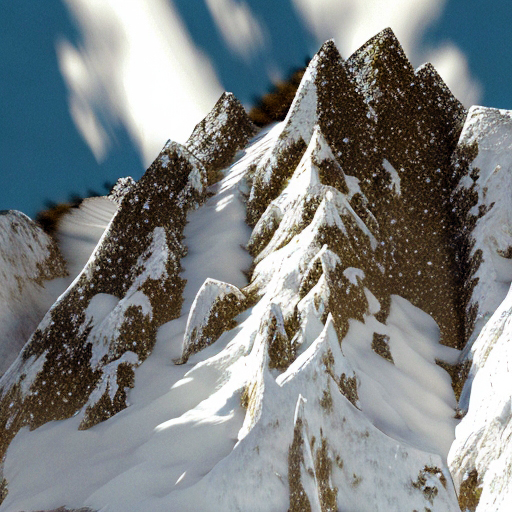}}};   
             \node (fig2) at (-6,-6)
               {\fbox{\includegraphics[width=0.08\textwidth]{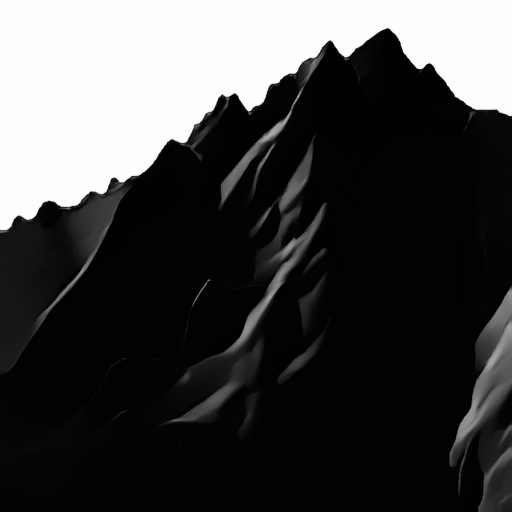}}};
        \end{tikzpicture} &
        \begin{tikzpicture}[every node/.style={anchor=north west,inner sep=0pt},x=1pt, y=-1pt,]  
             \node (fig1) at (0,0)
               {\fbox{\includegraphics[width=0.23\textwidth]{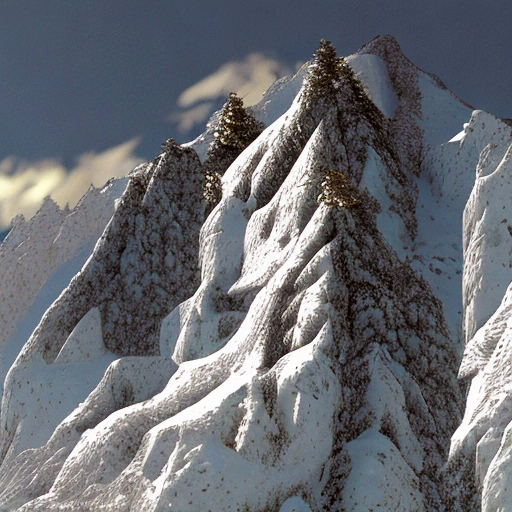}}};   
             \node (fig2) at (-6,-6)
               {\fbox{\includegraphics[width=0.08\textwidth]{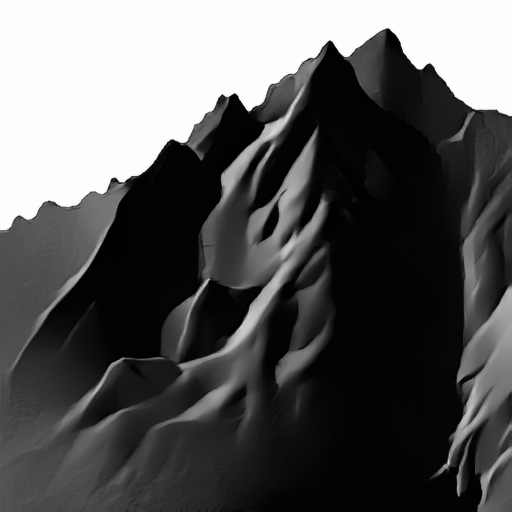}}};
        \end{tikzpicture} &
        \begin{tikzpicture}[every node/.style={anchor=north west,inner sep=0pt},x=1pt, y=-1pt,]  
             \node (fig1) at (0,0)
               {\fbox{\includegraphics[width=0.23\textwidth]{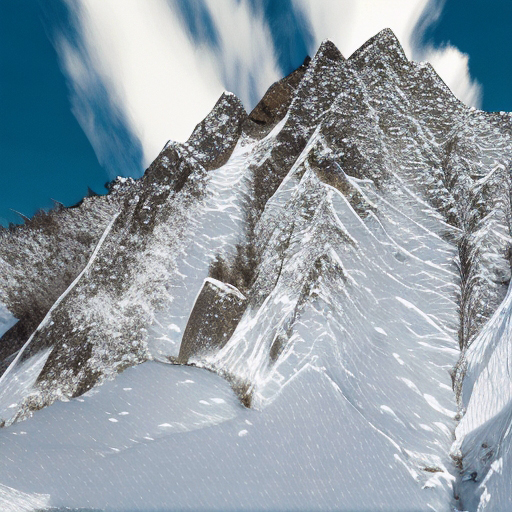}}};   
             \node (fig2) at (-6,-6)
               {\fbox{\includegraphics[width=0.08\textwidth]{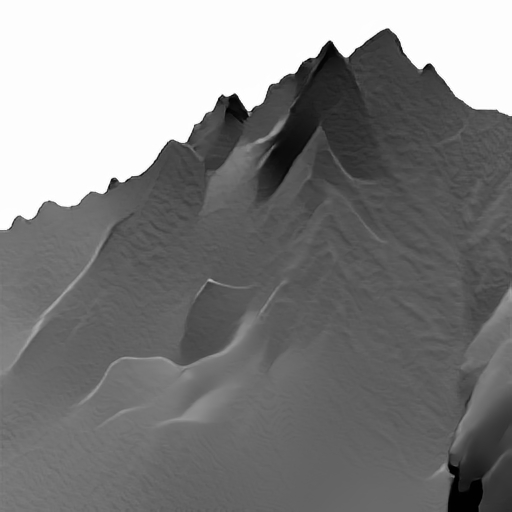}}};
        \end{tikzpicture} &
        \begin{tikzpicture}[every node/.style={anchor=north west,inner sep=0pt},x=1pt, y=-1pt,]  
             \node (fig1) at (0,0)
               {\fbox{\includegraphics[width=0.23\textwidth]{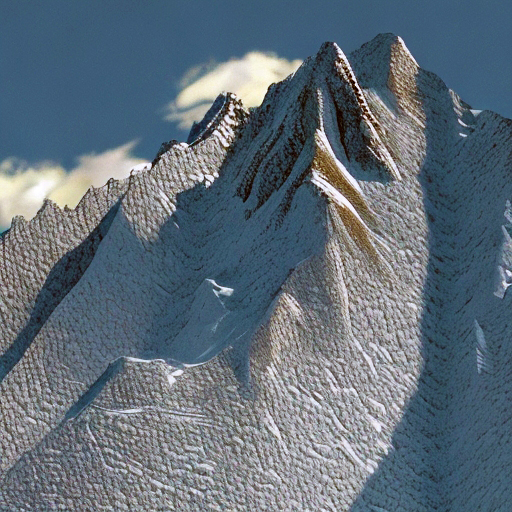}}};   
             \node (fig2) at (-6,-6)
               {\fbox{\includegraphics[width=0.08\textwidth]{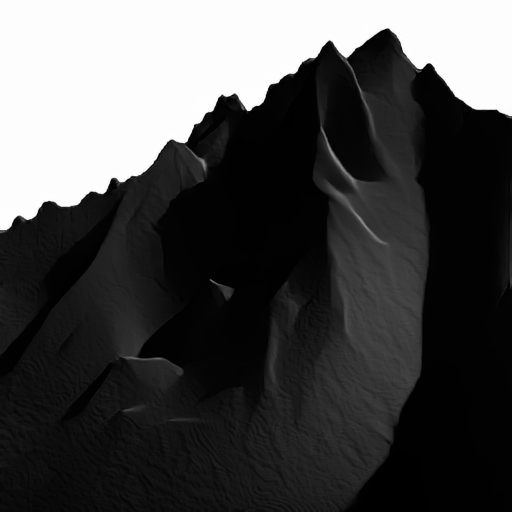}}};
        \end{tikzpicture}
        \\

        \rotatebox{90}{\textit{Image}} &
        \begin{tikzpicture}[every node/.style={anchor=north west,inner sep=0pt},x=1pt, y=-1pt,]  
             \node (fig1) at (0,0)
               {\fbox{\includegraphics[width=0.23\textwidth]{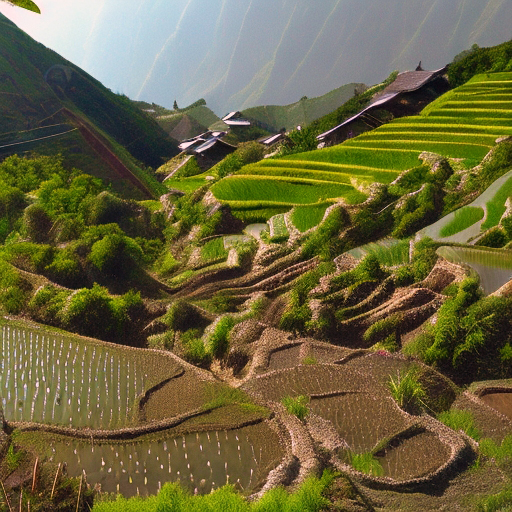}}};   
             \node (fig2) at (-6,-6)
               {\fbox{\includegraphics[width=0.08\textwidth]{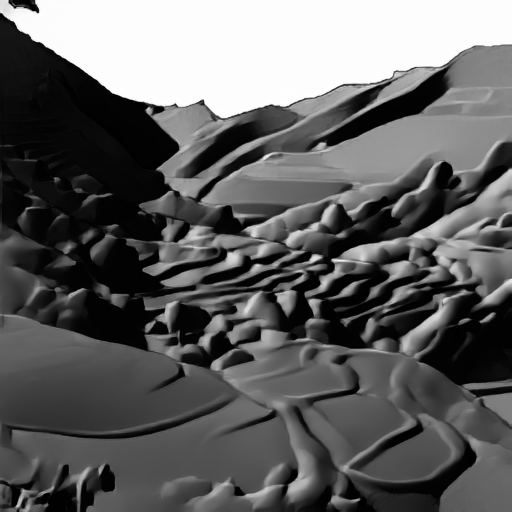}}};
        \end{tikzpicture} &
        \begin{tikzpicture}[every node/.style={anchor=north west,inner sep=0pt},x=1pt, y=-1pt,]  
             \node (fig1) at (0,0)
               {\fbox{\includegraphics[width=0.23\textwidth]{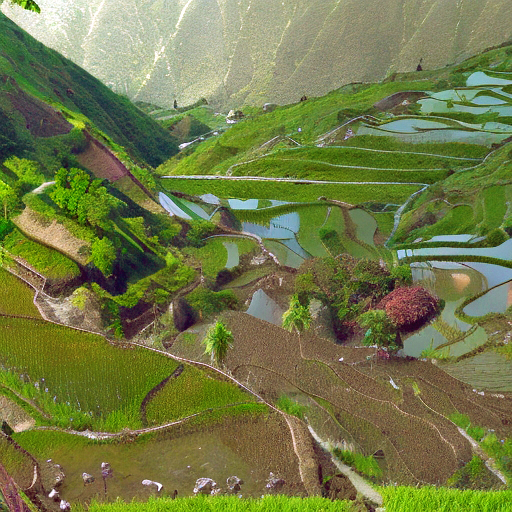}}};   
             \node (fig2) at (-6,-6)
               {\fbox{\includegraphics[width=0.08\textwidth]{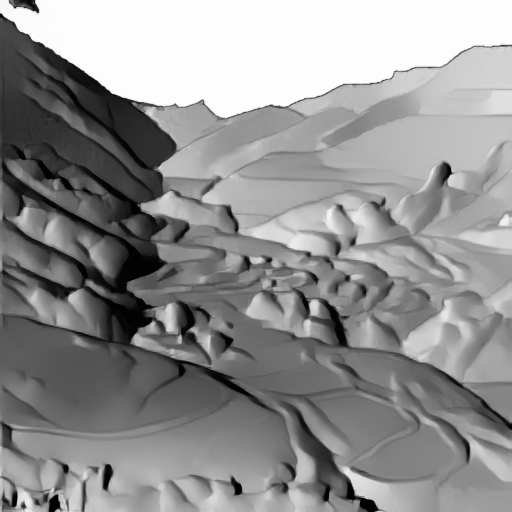}}};
        \end{tikzpicture} &
        \begin{tikzpicture}[every node/.style={anchor=north west,inner sep=0pt},x=1pt, y=-1pt,]  
             \node (fig1) at (0,0)
               {\fbox{\includegraphics[width=0.23\textwidth]{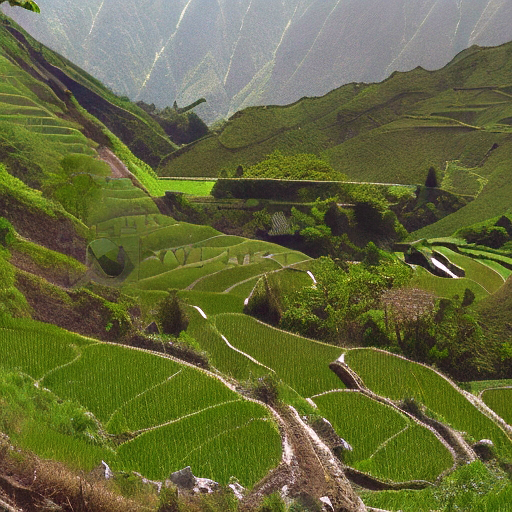}}};   
             \node (fig2) at (-6,-6)
               {\fbox{\includegraphics[width=0.08\textwidth]{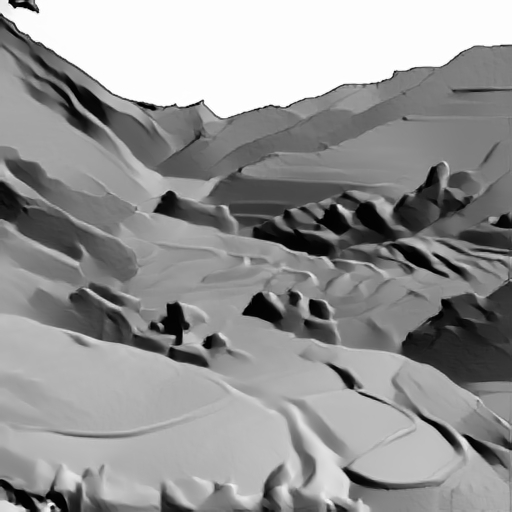}}};
        \end{tikzpicture} &
        \begin{tikzpicture}[every node/.style={anchor=north west,inner sep=0pt},x=1pt, y=-1pt,]  
             \node (fig1) at (0,0)
               {\fbox{\includegraphics[width=0.23\textwidth]{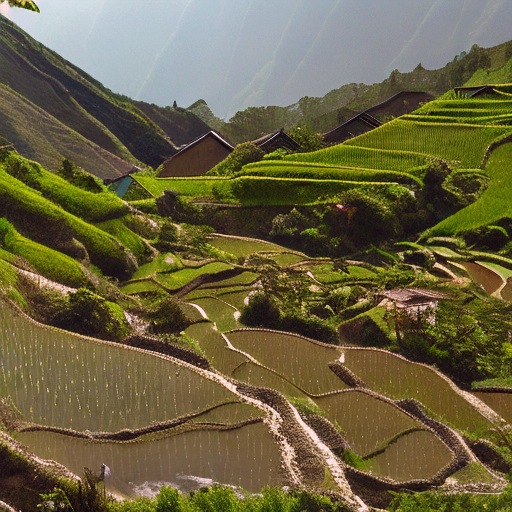}}};   
             \node (fig2) at (-6,-6)
               {\fbox{\includegraphics[width=0.08\textwidth]{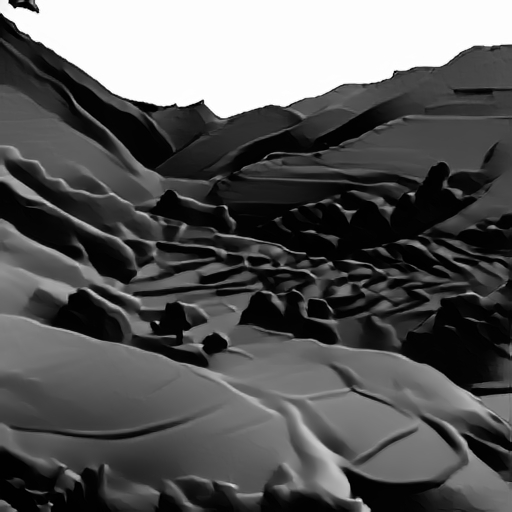}}};
        \end{tikzpicture}
        \\
         & Lighting 1 & Lighting 2 & Lighting 3 & Lighting 4
    \end{tabular}}

    \vspace{-8pt}
    \caption{\textbf{Out-of-Domain Image Synthesis with Controllable Lighting}. 
    Our method learns to control the generation process yet maintains the prior of SD \cite{rombach2022high}. 
    We show various scenes and styles under changing lighting conditions. 
    The first three images were obtained with estimated normal of real-world images, while for the last two, we used images generated with SDXL \cite{podell2023sdxl}. 
    See supplemental for details. 
    \readyForFeedback}
    \label{fig:exp:od_controllable_image_synthesis_id}
\end{figure*}
}

\newcommand{\TabLightingEvaluation}{
\begin{table}
\caption{\textbf{Perceptual Image Generation Quality.}
We perform a user study to assess the quality of the generated images. 
Since there exists no other method capable of explicit lighting control, we compare against SD \cite{rombach2022high}. 
The users are provided two images generated with SD \cite{rombach2022high} and our method using the same prompt. 
We report the perceptual quality regarding the image (I-PQ), lighting (L-PQ), and text alignment (T-PQ). 
Thanks to better lighting, our results are preferred in every aspect, not just in lighting. 
\readyForFeedback
}
\label{tab:exp:lighting_evaluation}
\begin{center}
\vspace{-9pt}
\begin{tabular}{l|ccc}
  \toprule
    & L-PQ $\uparrow$ & I-PQ $\uparrow$ & T-PQ $\uparrow$  \\
   \midrule 
  SD  & 4.43 & 39.14 & 44.14 \\
  Ours  & \textbf{95.57} & \textbf{60.86} & \textbf{55.86} \\
  \bottomrule
\end{tabular}
\vspace{-18pt}
\end{center}
\end{table}
}

\newcommand{\figAblationLightingRepresentation}{
\begin{figure}[t]
    \centering
    \setlength\tabcolsep{0.5pt}
    \resizebox{\columnwidth}{!}{
    \fboxsep=0pt
    \begin{tabular}{ccc}
        \fbox{\includegraphics[width=0.3\columnwidth]{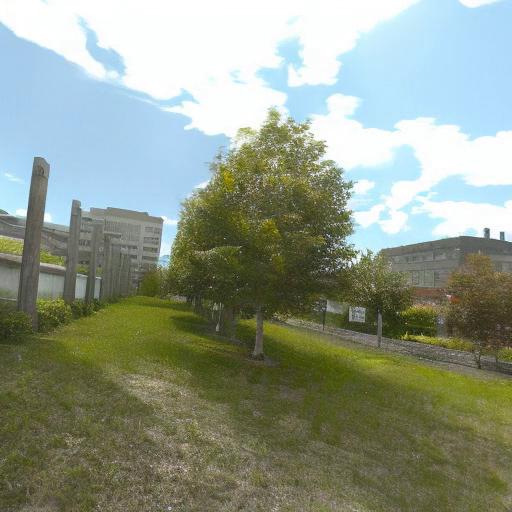}}
        & 
        \begin{tabular}[b]{cc}
            \raisebox{14pt}{\small{Direct}} & 
            \fbox{\includegraphics[width=0.1415\columnwidth]{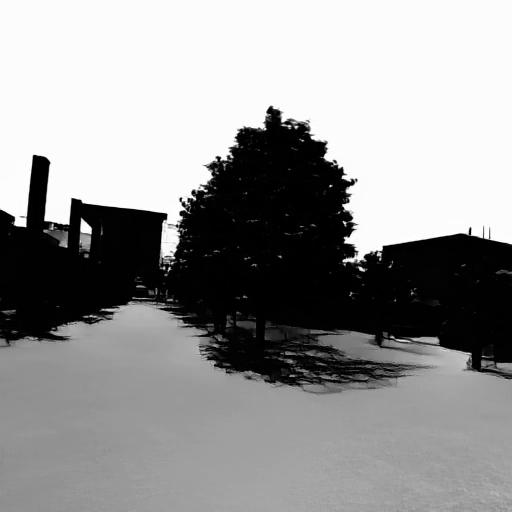}}\\
            \fbox{\includegraphics[width=0.1415\columnwidth]{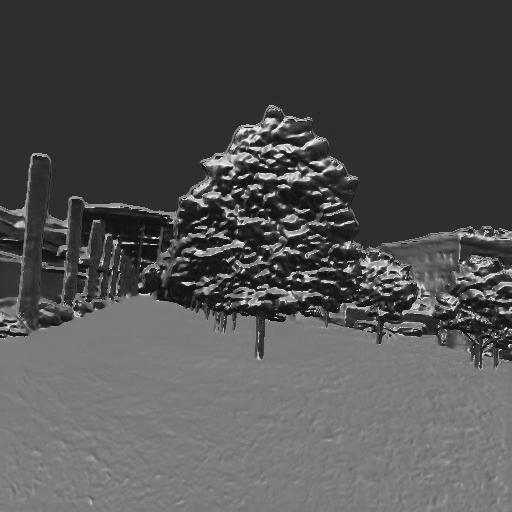}} &
            \raisebox{14pt}{\small{N-dot-L}} \\
        \end{tabular}
        & 
        \fbox{\includegraphics[width=0.3\columnwidth]{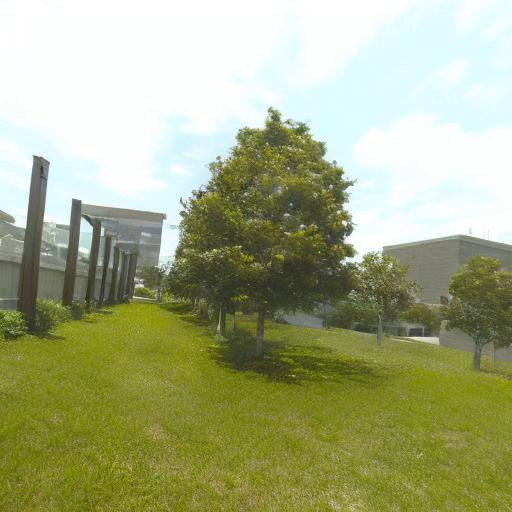}}
        \\
        w/ N-dot-L & \small{Input Shading} & Ours
    \end{tabular}}

    \vspace{-3pt}
    \caption{\textbf{Effect of Lighting Representation}. 
    We show that cast shadows provide essential information for the generation process. 
    We compare our method against a simple N-dot-L shading conditioning, which provides only coarse lighting information to the model, leading to inconsistent lighting with less control. 
    \readyForFeedback}
    \label{fig:exp:abl_lighting_representation}
    \vspace{-3pt}
\end{figure}
}

\newcommand{\figAblationNormalConditioning}{
\begin{figure}[t]
    \centering
    \setlength\tabcolsep{1.25pt}
    \resizebox{\columnwidth}{!}{
    \fboxsep=0pt
    \begin{tabular}{c|ccc}
        \begin{tabular}[b]{cc}
            \rotatebox{90}{Shading} & 
            \fbox{\includegraphics[width=0.19\columnwidth]{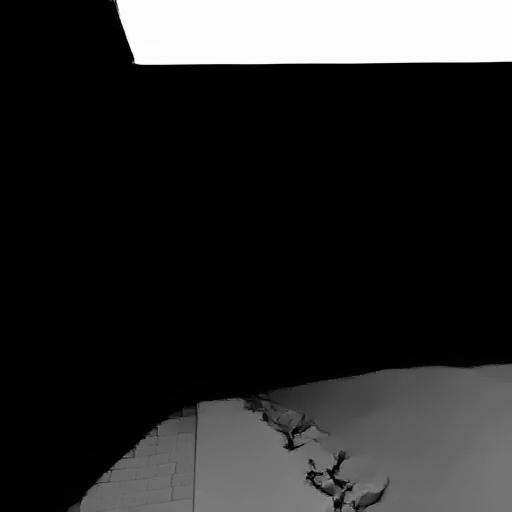}}\\
            \rotatebox{90}{Normal} &
            \fbox{\includegraphics[width=0.19\columnwidth]{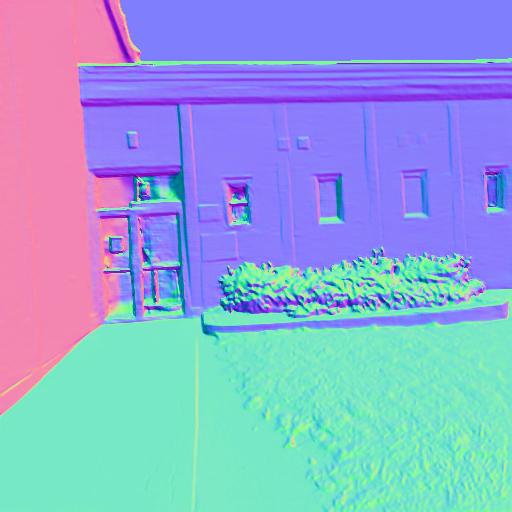}}\\
        \end{tabular}
        & 
        \fbox{\includegraphics[width=0.40\columnwidth]{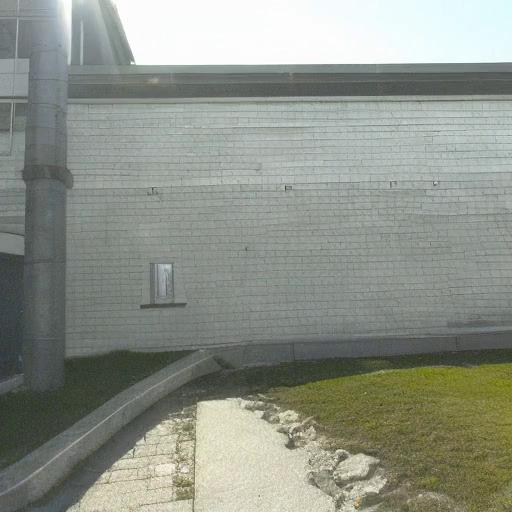}}
        & 
        \fbox{\includegraphics[width=0.40\columnwidth]{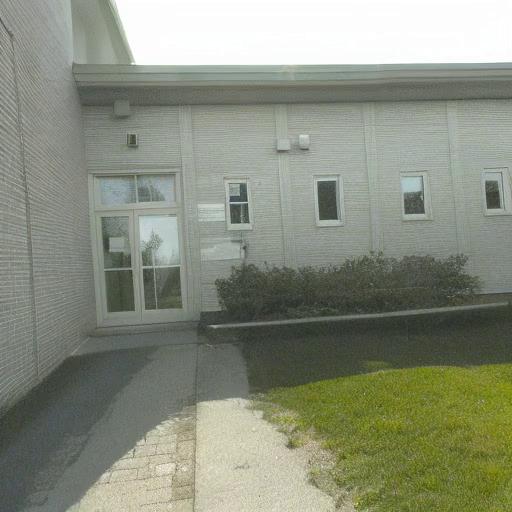}}
        \\
        & w/o Normals & Ours
    \end{tabular}}

    \vspace{-6pt}
    \caption{\textbf{Effect of Normal Conditioning}. 
    Without normal conditioning, it is impossible for the model to infer geometry in the shadowed regions. 
    \readyForFeedback}
    \label{fig:exp:abl_normal_conditioning}
    \vspace{-15pt}
\end{figure}
}

\newcommand{\TabControlConsistency}{
\begin{table}
\caption{\textbf{Control Consistency.}
We estimate the shading (L) and normal (N) of generated images on our test set and compare them against the control signal in image space (PSNR) and in angular error measured in degrees (AE). 
Conditioning on our direct shading (DS) achieves the best lighting quality; however, it does not ensure consistent normals (\cref{fig:exp:abl_normal_conditioning}).
Providing normals to the model helps with minimal cost of the lighting quality. 
\readyForFeedback
}
\label{tab:exp:control_consistency}
\begin{center}
\vspace{-3pt}
\resizebox{\columnwidth}{!}{
\begin{tabular}{l|cc|cc}
    \toprule
    & L-PSNR $\uparrow$ & L-AE $\downarrow$ & N-PSNR $\uparrow$ & N-AE $\downarrow$ \\
    \midrule 
    N-dot-L Shading  & 
    6.43 {\footnotesize $\pm$ 2.20} & 
    37.23 {\footnotesize $\pm$ 23.79} & 
    16.45 {\footnotesize $\pm$ 2.53} & 
    21.76 {\footnotesize $\pm$ 9.96} \\
    
    Direct Shading (DS) & 
    \textbf{13.04} {\footnotesize $\pm$ 3.57} & 
    \textbf{27.30} {\footnotesize $\pm$ 19.13} & 
    17.30 {\footnotesize $\pm$ 2.75} & 
    18.73 {\footnotesize $\pm$ 9.54} \\
    
    DS + Normals (Ours)  & 
    12.69 {\footnotesize $\pm$ 3.52} & 
    28.59 {\footnotesize $\pm$ 20.46} & 
    \textbf{17.47} {\footnotesize $\pm$ 2.72} & 
    \textbf{18.28} {\footnotesize $\pm$ 9.28} \\
    
    \bottomrule
\end{tabular}}
\vspace{-12pt}
\end{center}
\end{table}
}

\newcommand{\figRelighting}{
\begin{figure}[t]
    \centering
    \setlength\tabcolsep{0.5pt}
    \resizebox{\columnwidth}{!}{
    \fboxsep=0pt
    \begin{tabular}[c]{ccc}
        \raisebox{9pt}{
            \fbox{\includegraphics[width=0.25\columnwidth]{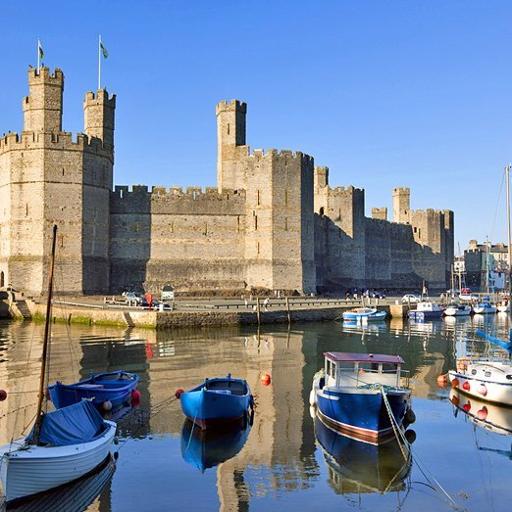}}
        }
        & 
        \fbox{\includegraphics[width=0.33\columnwidth]{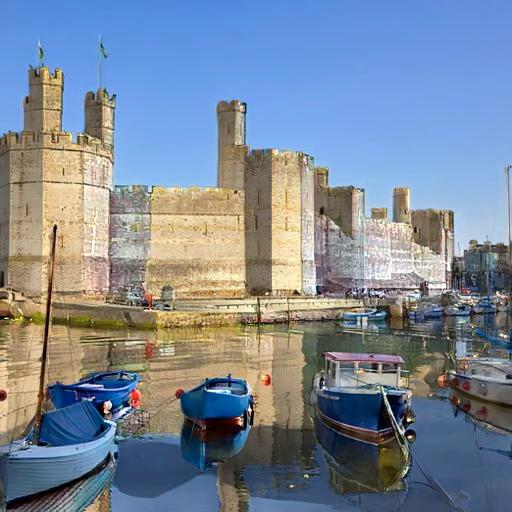}}
        & 
        \fbox{\includegraphics[width=0.33\columnwidth]{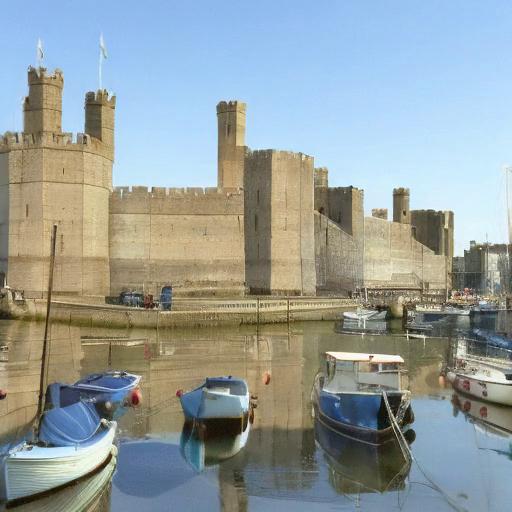}} \\
        
        \raisebox{9pt}{
            \fbox{\includegraphics[width=0.25\columnwidth]{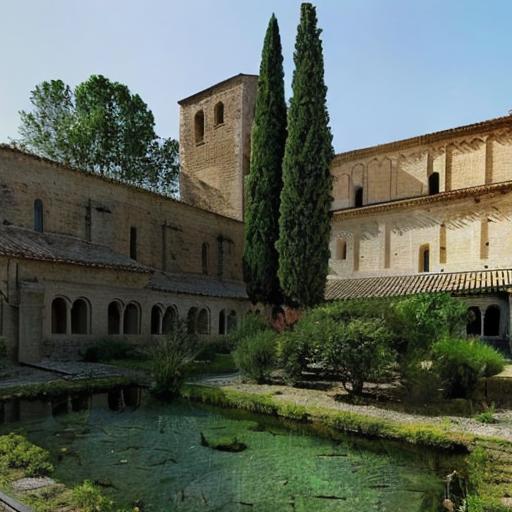}}
        }
        & 
        \fbox{\includegraphics[width=0.33\columnwidth]{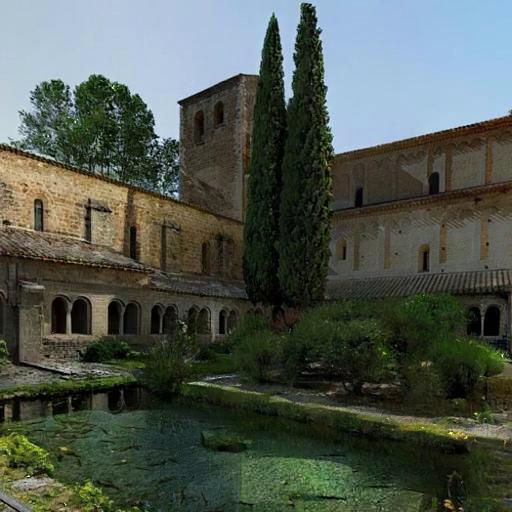}}
        & 
        \fbox{\includegraphics[width=0.33\columnwidth]{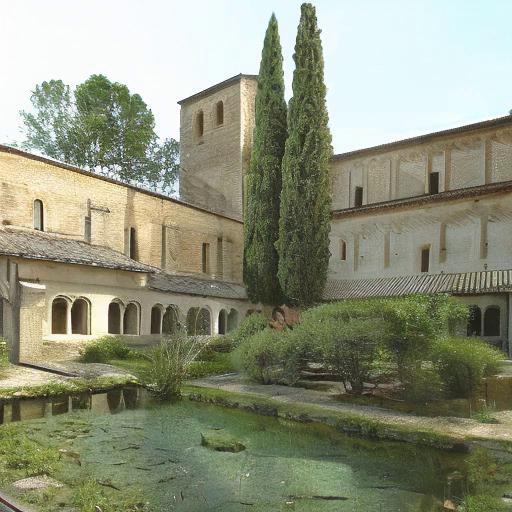}} \\
        
        Input & OutCast \cite{griffiths2022outcast} & Ours
    \end{tabular}}

    \vspace{-6pt}
    \caption{\textbf{Relighting of Real-World Images}. 
    We train a relighting network and evaluate it on real-world images.
    Utilizing the diffusion prior helps the generalization to real samples, especially for shading disambiguation. 
    \readyForFeedback}
    \label{fig:exp:relighting}
\end{figure}
}

\newcommand{\figIdentityPreservation}{
\begin{figure}[t]
    \centering
    \setlength\tabcolsep{1.25pt}
    \resizebox{\columnwidth}{!}{
    \fboxsep=0pt
    \begin{tabular}[t]{ccc}
        \fbox{\includegraphics[width=0.33\columnwidth]{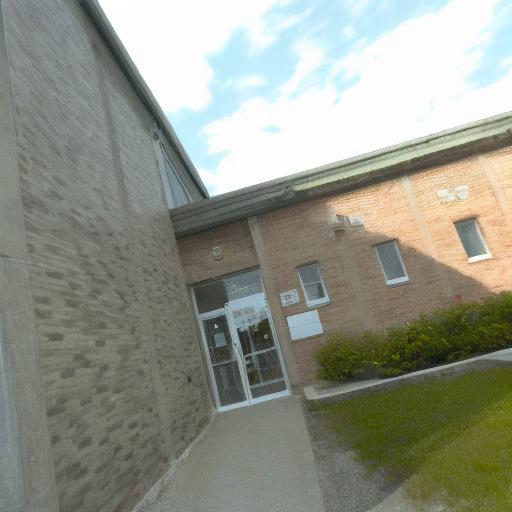}}
        & 
        \fbox{\includegraphics[width=0.33\columnwidth]{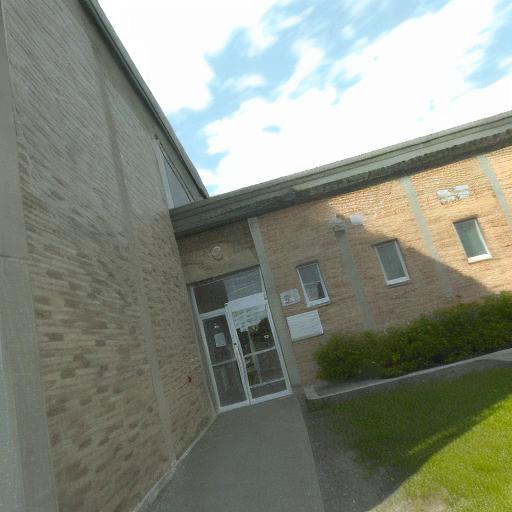}}
        & 
        \begin{tikzpicture}[every node/.style={anchor=north east,inner sep=0pt},x=-1pt, y=-1pt,]  
             \node (fig1) at (0,0)
               {\fbox{\includegraphics[width=0.33\columnwidth]{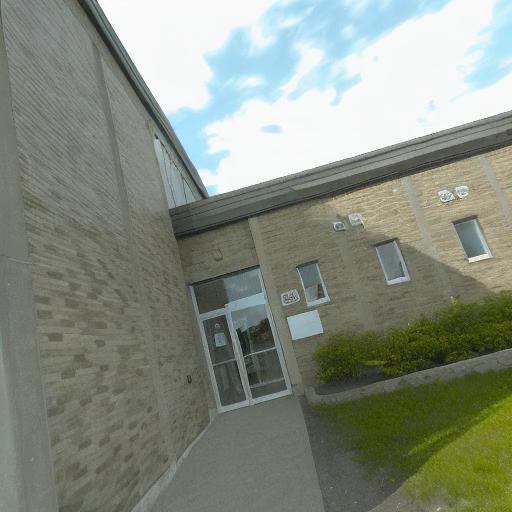}}};   
             \node (fig2) at (-18,-18)
               {\fbox{\includegraphics[width=0.08\textwidth]{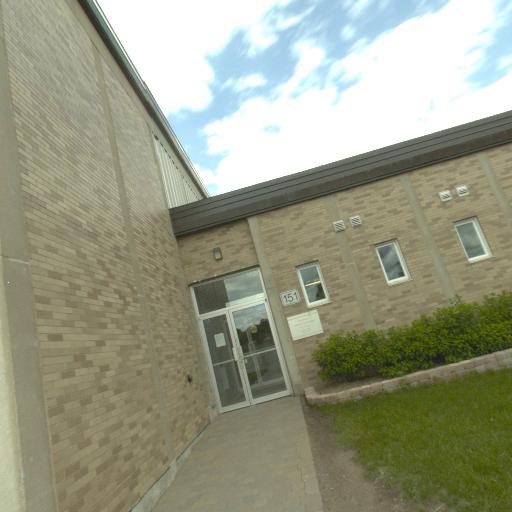}}};
            \node (fig3) at (25,-18)
               {Input};
        \end{tikzpicture} \\
        ControlNet \cite{zhang2023controlnet} & w/ Control Decoder & Ours
    \end{tabular}}

    \vspace{-6pt}
    \caption{\textbf{Identity Preservation}. 
    We ablate the effect of our control architecture on relighting. 
    ControlNet \cite{zhang2023controlnet} - left - is prone to ignoring part of the control signal, the wall turns reddish and the shadow gets softened. 
    Our Control Decoder - middle - with control reconstruction loss helps. 
    Our full residual architecture - right - takes another step and achieves high consistency. 
    \readyForFeedback}
    \label{fig:exp:abl_identity_preservation}
    \vspace{-9pt}
\end{figure}
}

\newcommand{\TabRelighting}{
\begin{table}
\caption{\textbf{Relighting Evaluation.}
We quantitatively compare our relighting method to OutCast on geometry (PSNR), image quality (FID, I-PQ) and lighting quality (L-PQ), where PQ refers to perceptual quality from our user study. 
The shadows are usually stronger for OutCast \cite{griffiths2022outcast}, leading to a slightly higher perceptual score. 
Our method achieves consistent relighting with more realistic image quality. 
\readyForFeedback
}
\label{tab:exp:relighting}
\vspace{-9pt}
\begin{center}
\resizebox{\columnwidth}{!}{
\begin{tabular}{l|cc|cc}
  \toprule
    & PSNR $\uparrow$ & FID $\downarrow$ & L-PQ $\uparrow$ & I-PQ $\uparrow$  \\
   \midrule 
  OutCast  & 19.79 {\footnotesize $\pm$ 4.39}  & 71.08 & \textbf{54.27} & 36.47 \\
  Ours - w/o RCD  & 20.24 {\footnotesize $\pm$ 3.29} & 76.28 & - & - \\
  Ours  & \textbf{20.44} {\footnotesize $\pm$ 3.34} & \textbf{64.18} & 45.74 & \textbf{63.53} \\
  \bottomrule
\end{tabular}}
\vspace{-18pt}
\end{center}
\end{table}
}

\def\paperID{15919} %
\def\confName{CVPR}
\def\confYear{2024}

\title{LightIt: Illumination Modeling and Control for Diffusion Models}

\author{Peter Kocsis$^{1}$\thanks{Research done during internship at Adobe.}\\
\and
Julien Philip$^{2}$\\
\and
Kalyan Sunkavalli$^{2}$\\
\and
Matthias Nie{\ss}ner$^{1}$\\
\and
Yannick \\ Hold-Geoffroy$^{2}$\\
\and
}

\begin{document}

\maketitle

\begin{abstract}\readyForFeedback
\vspace{-3pt}
We introduce LightIt, a method for explicit illumination control for image generation. 
Recent generative methods lack lighting control, which is crucial to numerous artistic aspects of image generation such as setting the overall mood or cinematic appearance. 
To overcome these limitations, we propose to condition the generation on shading and normal maps. 
We model the lighting with single bounce shading, which includes cast shadows.
We first train a shading estimation module to generate a dataset of real-world images and shading pairs. 
Then, we train a control network using the estimated shading and normals as input. 
Our method demonstrates high-quality image generation and lighting control in numerous scenes. 
Additionally, we use our generated dataset to train an identity-preserving relighting model, conditioned on an image and a target shading. 
Our method is the first that enables the generation of images with controllable, consistent lighting and performs on par with specialized relighting state-of-the-art methods. 

\end{abstract}

\section{Introduction\readyForFeedback}
\label{sec:introduction}
\vspace{-6pt}
Generative imaging has significantly evolved over recent years. 
Diffusion models have shown outstanding capabilities to learn strong priors on large-scale real-image datasets.
When used in conjunction with joint language-image embeddings, they have been successfully used for high-quality text-driven image generation. 
However, as these methods do not model light transport explicitly, the lighting in the produced images is uncontrollable and often inconsistent \cite{guo2023shadowdiffusion,corvi2023detection}. 
However, lighting is an essential component of most images; without explicit control, artists rely on tedious and uncertain prompt engineering to try to control it. 

Recent methods \cite{zhang2023controlnet,mou2023t2i,Huang2023composer} have introduced control for various aspects of the generated images, for instance, depth or normals can be used to guide the geometry. 
Fine-grained control over the placement of generated objects can already be achieved \cite{mou2023dragon,shi2023dragdiffusion} and diffusion inversion enables modification of generated images \cite{song2020denoising,wallace2023edict}. 
However, none of the methods can provide consistent and explicitly controllable lighting, which is the essence of photo-realistic images. 

Diffusion models achieve incredible performance when trained on large datasets.
However, the lack of real-world lighting datasets is a major obstacle hindering progress on lighting control. 
Obtaining the lighting of a scene is a time-consuming task, requiring the decomposition of its appearance into lighting and material properties. 
However, we hypothesize that diffusion models do not require fine-grained lighting information, thus simplifying the required decomposition. 
Our analysis demonstrates that estimated single-bounce shading maps provide sufficient information and can be automatically obtained from real-world images. 

In this work, we propose a single-view shading estimation method to generate a paired image and shading dataset. 
Given a single input image---either captured or generated---our model predicts a 3D density field, in which we trace rays toward a light to obtain cast shadows. 
Together with the estimated normals, giving us the cosine term, we predict single-bounce shading maps. 
This method notably allows us to generate shading maps for arbitrary lighting directions from a single image. 
Given outdoor panoramas, from which we can obtain the light direction, we thus generate a paired dataset of images and their shading. 
This dataset enables us to provide lighting control to the image generation process, which we also condition on normals to guide geometry.

As an additional application of our proposed illumination model, we further propose a relighting module conditioned on an input image and a target shading. 
Thanks to the strong natural image prior of Stable Diffusion (SD) \cite{rombach2022high}, we obtain better generalization to real-world samples compared to methods from the literature trained on synthetic data. 
In summary, our main contributions are:
\begin{itemize}
\item 
We generate a paired image-shading dataset using our single-view, density field-based lighting estimation model, enabling single-bounce shading estimation for arbitrary lighting directions. 
\item 
We introduce lighting conditioning for controllable and coherent image generation using a diffusion-based model. 
\item 
We propose an identity-preserving relighting diffusion module utilizing the image prior for better generalization. 
\end{itemize}

\figShadingEstimation

\section{Related Work\incomplete}
\label{sec:relatedwork}

\noindent\textbf{Lighting controlled image generation. }
Generative imaging is a recent field that started to receive attention with the invention of GANs \cite{goodfellow2014generative,radford2015unsupervised}. 
However, one of the main challenges is the lack of control over the image generation process. 
To address this issue, methods such as StyleGAN \cite{karras2019style} have been developed to provide control handles. 

Recently, diffusion-based models were proposed to perform generative imaging \cite{rombach2022high,ho2020denoising,song2020denoising,po2023diffusion}, enabling photoreal outputs from text prompts and democratizing generative imaging to the masses. 
As artists experimented with this new tool and explored its capabilities \cite{zhan2023does}, the need for more control over the generation process arose. 
In particular, ControlNet \cite{zhang2023controlnet} and T2I-Adapter \cite{mou2023t2i} were recently proposed to allow users to control the generated image using a variety of modalities at the cost of image quality \cite{ku2023imagenhub}. 
However, no approach exists for explicitly controlling the lighting of the generated imagery.

\noindent\textbf{Relighting. }
Image relighting has traditionally been performed using classical approaches such as image-based rendering \cite{debevec2000acquiring} or shape from shading \cite{barron2014shape}. 
The emergence of deep learning brought novel relighting approaches, initially using style transfer \cite{gatys2016image,luan2017deep,jing2019neural} or image-to-image translation \cite{isola2017image,zhou2019deep,sun2019single}. 
Specialized relighting methods have begun with \citet{xu2018deep} that learns a relighting function from five images captured under predetermined illumination. \citet{sengupta2021light} proposes to replace the traditional acquisition techniques with a regular monitor and camera setup. 
We encourage the reader to read the excellent review in \cite{tewari2020state} about rendering-based relighting. 

Closer to our work, scene relighting methods both multi-view \cite{philip2019multi,philip2021free,nicolet2020repurposing} and single-view \cite{yu2020self,griffiths2022outcast} generally use a combination of geometric and shading priors with a neural network to produce relit results. 
Outdoor NeRF-based relighting methods \cite{rudnev2022nerf,wang2023fegr} have been recently proposed, bringing the power of this implicit volumetric representation to relighting. 
Close in spirit to our shading model, OutCast \cite{griffiths2022outcast} proposes to use depth and a large 3D CNN to process depth features sampled in image space to implicitly predict ray intersection. 
Our method builds on several of these ideas to propose a scene relighting approach combining volumetric scene representation, and explicit shadow ray-marching with diffusion-based image generation.

\section{Method\readyForFeedback}
\label{sec:method}
Our method adds lighting control to the image generation process of a diffusion-based model. 
We develop a shading estimation method (\cref{sec:method:shading}) and generate a dataset of paired real images and shading maps (\cref{sec:method:dataset}) to train a control module for SD \cite{rombach2022high} (\cref{sec:method:diffusion}). 
Our dataset enables additional applications, such as relighting (\cref{sec:method:relighting}). 

\subsection{Shading estimation}
\label{sec:method:shading}
To control the illumination of generated images, we aim to provide lighting information to the diffusion model. 
Estimating the shading from a single image is a challenging task even in the presence of depth estimation \cite{griffiths2022outcast}. 
Inspired by Outcast \cite{griffiths2022outcast}, we develop a lightweight model to estimate \emph{direct shading}, i.e. single-bounce illumination, from a single input image, which provides information about both shading and cast shadows.
We show this pipeline in \cref{fig:method:shading_estimation}. 

Specifically, we train a shading estimation model, which takes an image, a light direction, and a solid angle as input.
A small 2D CNN (FeatureNet) first encodes the image to obtain a set of features.
Then, using a pre-trained depth estimator (\cref{sec:normal_map}), we unproject these features to a multi-plane representation in Normalized Device Coordinates (NDC). 
Given a pixel's depth, the features are linearly distributed between the two planes closest to the depth. 

A small 3D CNN (DensityNet) processes the multi-plane and predicts a 3D density field. 
We render a cast shadow map for the light direction and angle using volumetric ray-marching. 
Finally, a 2D CNN (ShadingNet) transforms this shadow map and an N-dot-L cosine term map into coarse direct shading. 
To further improve the shading estimation quality, we apply a refinement module, which uses the input image and the predicted coarse shading. 

We train our model on synthetic pairs of rendered images and shadings using an $l_2$ loss. 
To better guide the training, we add an $l_2$ loss on the predicted depth and the expected depth of the density field from the camera.
When only a depth map is available, we use the N-dot-L shading image instead of the RGB image, which our method is robust to.

\subsection{Lighting-conditioned diffusion}
\label{sec:method:diffusion}
Our main goal is to provide explicit lighting control to a pre-trained diffusion model. 
Inspired by ControlNet \cite{zhang2023controlnet}, we train an additional module that provides control signals to the intermediate features of SD \cite{rombach2022high}, as depicted in \cref{fig:method:training_pipeline}. 

We use lighting information represented as direct shading maps as conditioning, which we concatenate to the normal map to provide geometric information to the model. 
Similarly to ControlNet \cite{zhang2023controlnet}, our control modules contain zero convolutions to introduce the control gradually. 

During training, we keep SD \cite{rombach2022high} fixed and optimize only our control module consisting of a Residual Control Encoder and Decoder (RCE, RCD) and a Lighting Control network. 
We found that using the architecture of ControlNet \cite{zhang2023controlnet} is prone to ignoring part of the control signal. 
Indeed as mentioned in the original paper, the controls tend to be picked up suddenly. 
We believe this might be due to low gradients early on as the encoder does not provide a meaningful signal to the control module.
To avoid this issue, we develop a more stable encoder module, RCE, and next to the diffusion noise prediction loss we additionally supervise the training with an $l_2$ loss on the control reconstruction obtained with our RCD. 
RCE and RCD use residual blocks for more stable control flow and the reconstruction loss ensures that the full control signal is provided to the light control module. 
During inference, we do not need the RCD.

\figTrainingPipeline

\subsection{Relighting Application}
\label{sec:method:relighting}
Besides controllable image generation, our lighting representation and dataset can also be employed for relighting applications. 
Relighting methods are usually trained on synthetic data leading to domain gap or on limited paired real data \cite{tewari2020state}.
Adding relighting capability to pre-trained diffusion models opens up a novel way of utilizing pre-trained image priors. 
To achieve this, we propose to condition the generation on an input image and a target shading as opposed to normals and shading for the generation task.

\noindent\textbf{Dataset. }
To avoid training on synthetic renderings leading to domain gap, we use predicted relit images (\cref{fig:method:dataset_pipeline}). 
Given cropped images and random lighting conditions, we use our shading estimation method to generate target coarse shading maps and predict relit images with OutCast \cite{griffiths2022outcast}.

\noindent\textbf{Training. }
To avoid inheriting the artifacts of OutCast, we use the relit images as input and target the real image as output. 
This way, our output domain is real-world images, and our model is able to utilize the strong prior of SD \cite{rombach2022high}.

\subsection{Dataset}
\label{sec:method:dataset}
We use the Outdoor Laval Dataset \cite{hold2019deep}, which consists of real-world HDR panoramas encoded as a latitude-longitude map. 
Given the full panorama, we determine the Sun's direction by detecting the brightest pixel in the panorama image and transforming it to an angular direction, which helps our shading estimation. 
Our dataset contains $51250$ samples of LDR images and text prompts with corresponding estimated normal and shading maps, as shown in \cref{fig:method:dataset_pipeline}. 

\noindent\textbf{Image}
We crop $250$ images of resolution $512\times512$ from each panorama. 
For each image, we use randomized camera parameters with varying field-of-view, elevation, and roll angles, as described in our supplemental. 
We normalize the images to have $0.5$ mean intensity.

\noindent\textbf{Depth and Normal}
\label{sec:normal_map}
As a first step in our dataset generation, we estimate the per-pixel surface normals of each image. 
To this end, we use the same depth estimator as OutCast \cite{griffiths2022outcast}, dubbed DepthNet. 
In summary, it is a segformer-based depth estimator \cite{xie2021segformer} trained on the datasets proposed in \cite{vasiljevic2019diode,kornilova2022smartportraits,roberts2021hypersim,wang2021irs,Silberman:ECCV12,li2018megadepth,garg2019learning}. The model was trained using the loss function proposed in \cite{ummenhofer2017demon}. We project this estimated depth to a point cloud using $x_i = \frac{1}{f} u_i z_i$, where $u_i$ is the pixel's image coordinates, $f$ is the focal length in pixels, and $z_i$ is the estimated depth at pixel $i$. We perform the same operation on $y_i$ to obtain $\vec{\mathbf{p}}_i = [x_i, y_i, z_i]$. Finally, we obtain the per-pixel normal by computing the discrete derivative over the point cloud, as 
\begin{equation}
    \vec{\mathbf{n}} = \frac{ \partial\vec{\mathbf{p}}}{\partial x} \times \frac{ \partial\vec{\mathbf{p}}}{\partial y} \;.
\end{equation}
We experimentally compared this approach to directly estimating surface normals, and the former provided a more robust estimation. 
We hypothesize that the larger corpus of publicly available depth maps datasets yields a better depth estimator, explaining this improved performance. 

At test time, we experimented with swapping our depth estimator with MiDaS \cite{ranftl2020towards,birkl2023midas} and achieved similar image generation results, when a plausible depth map is obtained. 

\noindent\textbf{Shading}
\label{sec:shading}
A simple lighting representation is to employ the depth map as geometry and determine the N-dot-L shading. 
While this conceptually simple representation is straightforward to implement from a depth map alone, it does not consider cast shadows. 
Thus, we use our shading estimation method (\cref{sec:method:shading}) to obtain refined direct shading maps. 

\noindent\textbf{Prompt}
To maintain the textual capabilities of our model, we include text prompts for each sample in our dataset. 
We use BLIP-2 \cite{li2023blip} to automatically caption the images. 

\figDatasetPipeline
\figConsistentImageSynthesis

\section{Experiments\incomplete}
\label{sec:experiments}

\subsection{Image Synthesis\readyForFeedback}
\label{sec:experiments:synthesis}

For image synthesis, our inputs consist of a shading map (\cref{sec:shading}) and optionally a normal map (\cref{sec:normal_map}). 
These maps can be estimated from either a guidance image or any text-to-image pipeline. 
\revise{Novel shading can be rendered using the coarse shading of our shading estimation}. 
All evaluations are performed on real images from our test set or in-the-wild images, never seen in training. 
We want to emphasize that, to the best of our knowledge, our method is the first to achieve this degree of lighting control on diffusion-based generative imaging. 

\noindent\textbf{Training.} 
We optimize our control module with the AdamW \cite{loshchilov2017decoupled} optimizer for two epochs using a learning rate of $1e\text{-}5$ with a control reconstruction weight of $1$.

\noindent\textbf{Inference.} We employ the DDPM sampler \cite{ho2020denoising} for $1000$ steps for quantitative and in-domain queries.
For custom text prompts and styles, we use DDIM sampler \cite{song2020denoising} with $100$ steps and a guidance scale of $7$ and early control stopping at timestep $200$ to avoid overruling the text guidance.

\vspace{-9pt}
\subsubsection{Lighting Consistency. }
\label{sec:experiments:synthesis:consistency}
We first qualitatively evaluate our method's capability to produce the desired lighting across various text prompts, provided in \cref{fig:exp:consistent_image_synthesis}. 
As can be seen, our method produces consistent, convincing shading across various styles following the target shading well. 

We also evaluate the lighting consistency of our model with a user study on $42$ participants shown in \cref{tab:exp:lighting_evaluation}. 
The study contains images both from our test set and in the wild using the predicted text prompt from our dataset and also manually prepared ones. 
The users are presented with the input maps along with generated images of our method and SD \cite{rombach2022high} and are asked to answer three questions: 1) which image corresponds the best to the lighting input, 2) which image matches best the input text prompt, and 3) which image has the best overall quality. 
This user study reveals that our method not only follows the desired lighting well but is also preferred more in terms of image quality and textural alignment. 
Lighting is an essential part of the perceptual image quality. 
SD \cite{rombach2022high} is not enforced to produce physically consistent lighting, leading to perceptually degraded images compared to ours.

\TabLightingEvaluation
\figIDControllableImageSynthesis
\figODControllableImageSynthesis

\subsubsection{Lighting Controllability\readyForFeedback}
\label{sec:experiments:synthesis:controllability}

In \cref{fig:exp:id_controllable_image_synthesis}, we show examples of novel lighting on images from our test set. 
Our results correctly follow the user-defined lighting (insets) shown in the surface shading. 

In \cref{fig:exp:od_controllable_image_synthesis_id}, we show examples of novel image generation with controlled lighting on in-the-wild inputs. 
Specifically, the normal and shading maps of the top three rows come from images taken from the internet\footnote{We obtained the licenses for their use.}, while the two bottom rows were entirely generated from a text prompt using Stable Diffusion. 
We then ran our normal estimation and coarse shading estimation approach on each image entirely automatically. 
In this setup, each lighting was generated independently, without care for identity preservation; only the initial noise was fixed to mitigate discrepancies. 

\figRelighting
\TabRelighting
\figIdentityPreservation

\subsection{Image Relighting\readyForFeedback}
\label{sec:experiments:relighting}
Our lighting representation and proposed dataset enable additional lighting-related applications, such as relighting. 
For this task, we use an input image and a target shading map as conditioning. 
We compare against our reimplementation of OutCast \cite{griffiths2022outcast} using our shading estimation. 
\todo{be careful to make sure it's not proposed as the main contribution.} 

\noindent\textbf{Training. } 
We use our extended dataset (\cref{sec:method:relighting}) and use the OutCast \cite{griffiths2022outcast} relit image with the source shading as conditioning and the original image as target. 
We train this model for six epochs using the process described in \cref{sec:experiments:synthesis}. 

\noindent{}\textbf{Inference. } 
We use the DDPM sampler with $1000$ steps and produce the text prompts automatically using BLIP-2 \cite{li2023blip}. 

\noindent{}\textbf{Evaluation. }
We first validate our relighting model qualitatively on in the wild images compared against OutCast \cite{griffiths2022outcast} in \cref{fig:exp:relighting}. 
OutCast provides very competitive results; however, being trained on synthetic data limits its generalization to real data.
Regions originally in shadow notably suffer from noise amplification. 
In contrast, our diffusion-based model provides visually pleasing results. 

We evaluate quantitatively in \cref{tab:exp:relighting}. 
We perform a cycle relighting experiment on our test set and predict the original real image from OutCast \cite{griffiths2022outcast} relit images (PSNR). 
Our method already outperforms OutCast without our proposed RCD module (\cref{sec:method:diffusion}), thanks to the diffusion prior. 
In addition, our full method produces more natural images with the best FID score \cite{heusel2017gans}. 
We further evaluate the perceptual quality (PQ) with a user study performed by $17$ people. 
The users were asked to evaluate the lighting consistency (L-PQ) and the overall image quality and realism (I-PQ) between our method and OutCast. 
OutCast usually provides stronger shadows that are perceptually slightly more consistent.
However, aligned with the FID, our method produces significantly more realistic results according to the users.

\subsection{Ablations\readyForFeedback}

\subsubsection{Image Synthesis\readyForFeedback}
\label{sec:experiments:synthesis:ablation}
\noindent\textbf{Does the model need cast shadows? }
Our key design choice is to provide information about cast shadows to the model. 
We argue that simpler lighting representation is insufficient because the latent space of a pre-trained diffusion model does not encode consistent global illumination. 
We qualitatively compare against a similar but simpler setup in \cref{fig:exp:abl_lighting_representation}, where the conditioning is an N-dot-L shading map without any cast shadow. 
Notably, our model can infer the overall lighting also with N-dot-L shading but fails to generate realistic shadows. 
In contrast, using direct shading offers much more appealing results with fine-grained shadow control. 
We provide more examples in our supplemental.

\vspace{18pt}
We provide quantitative results in \cref{tab:exp:control_consistency}.
We consider the estimated shading quality (L-PSNR) and the angular error between dominant light directions (L-AE).
Using direct shading outperforms N-dot-L with a high margin.

\TabControlConsistency
\noindent\textbf{Does the model need normals? }
Although most of the geometry can be inferred from our direct shading map, shadow regions do not provide any signal.
When the incident light is away from a surface normal $\cos (\vec{\mathbf{n}} \cdot \vec{\mathbf{l}} ) \! \le{} \! 0$ (attached shadow) or the light is occluded by some geometry (cast shadow) results in a uniform region of null values. 
Without additional geometric information, the model generates random geometric detail in those regions. 
We showcase this in \cref{fig:exp:abl_normal_conditioning}, where the model without normal generates a flat wall devoid of features in the shadow region, while our method correctly generates the expected door and windows. 

We quantitatively evaluate the effect of normal conditioning on our test set in \cref{tab:exp:control_consistency}. 
Using normals improves the normal consistency in the shadow regions.

\subsubsection{Image Relighting\readyForFeedback}
\label{sec:experiments:relighting:consistency}
\noindent\textbf{Does our architecture help identity preservation?}
Identity preservation is a crucial aspect of image relighting. 
Unfortunately, we have witnessed that diffusion-based image editing generally exhibits issues in identity preservation, especially in reproducing colors. 
Directly training ControlNet \cite{zhang2023controlnet} on our task produces changes in wall tint, for example, as shown in \cref{fig:exp:abl_identity_preservation}. 
We hypothesize that information pertaining to identity is lost in the encoder. 
\revise{We introduce our RCE and RCD modules (\cref{sec:method:diffusion}) to ensure enough control information is preserved, such as color during relighting }
which ensures that the feature map injected into the denoising U-Net keeps information to maintain the control signal.

\cref{tab:exp:relighting} quantitatively showcases the importance of our RCD (\cref{sec:method:diffusion}), where the image quality increases to an FID of 64.18 when our method is used without it. 

\subsection{Limitations and Future Work}
Our work assumes directional lighting, which is suitable for outdoor scenes.
However, our shading estimation method enables tracing rays in arbitrary directions.
Adapting our method to point and other light sources is an exciting avenue for future research.
Furthermore, our shading estimation requires lighting direction for the best results. 
Combining our method with a robust lighting estimation would allow training on much larger datasets.

\figAblationLightingRepresentation
\figAblationNormalConditioning

\section{Conclusion\readyForFeedback}
\label{sec:conclusion}

Recent diffusion-based generative imaging techniques have shown impressive text-to-image capabilities, producing breathtaking images on a whim. 
However, their controllability is limited, and adjusting important details such as lighting requires careful prompt engineering. 
In this work, we present a novel approach to explicitly control the illumination of images generated by a diffusion model. 
Our approach uses our direct shading representation, which contains both shading and shadow information. 
The shading map can be automatically computed from an existing picture or a generated image. 
Our method achieves high-quality results compared to existing methods while maintaining user-defined lighting. 
We believe that our method paves the way to increase the editability of diffusion-based generative imaging approaches.

\vspace{3pt}
\noindent\textbf{Acknowledgements.}
This work was supported by the ERC Starting Grant Scan2CAD (804724), the German Research Foundation (DFG) Grant ``Making Machine Learning on Static and Dynamic 3D Data Practical'', and the German Research Foundation (DFG) Research Unit ``Learning and Simulation in Visual Computing''. 

{
    \small
    \bibliographystyle{style/ieeenat_fullname}
    \bibliography{sections/bibliography}
}

\maketitlesupplementary
\appendix
\setcounter{page}{1}

In this supplementary material, first, we provide additional details on our method in \cref{sec:supp:method}, and on our experiments in \cref{sec:supp:experiment}. Finally, we show additional results in \cref{sec:supp:results}.

\section{Method Details}
\label{sec:supp:method}
We provide additional details about our method in the following sections. 

\subsection{Dataset Generation}
\label{sec:supp:method:dataset}
\noindent\textbf{Image. } 
Our dataset is generated from the Outdoor Laval dataset \cite{hold2019deep} consisting of $205$ HDR panorama images. 
Having access to the full panorama image gives us beneficial information about the dominant light direction. 

We render $250$ images from virtual cameras for every panorama. 
We use varying intrinsic and extrinsic parameters for the views with the constraints shown in \cref{tab:cupp:method:dataset:cropping}.
We normalize the images to have a mean intensity of $0.5$. 
We convert the images to LDR format by transforming to sRGB space using gamma correction ($\gamma=2.2$) and clipping to the $[0,1)$ range.

\subsection{Lighting Control}
\label{sec:supp:method:model}
Our full control module consists of three main modules: Residual Control Encoder (RCE), Lighting Control (LC) used during both training and inference; Residual Control Decoder (RCD) used only during training. 
RCE has approximately $2.9$M, LC $363$M, and RCD $1.3$M parameters.

\vspace{6pt}
\noindent\textbf{RCE. } 
Our RCE consists of $7$ residual blocks and maps from $512\times512\times3$ to $64\times64\times320$. 
The residual blocks follow the architecture of the ones in the diffusion UNet of \cite{rombach2022high}. 
Following \cite{zhang2023controlnet}, we use zero convolution at the beginning of our RCE. 

\vspace{6pt}
\noindent\textbf{RCD. }
During training, we use a decoder to reconstruct the control signal from the latent representation and use a reconstruction supervision to ensure most of the signal is preserved. 
Our RCD significantly improves the control consistency. 
Similar to our RCE, RCD consists of $7$ residual blocks and uses the same architecture as RCE, just in transposed order. 

\vspace{6pt}
\noindent\textbf{LC. }
Following \cite{zhang2023controlnet}, our LC uses the same architecture as the encoder of the UNet of \cite{rombach2022high}. 
LC takes the encoded control signal and returns the intermediate and encoded diffusion features to control the diffusion process \cite{zhang2023controlnet}.

\section{Experiment Details}
\label{sec:supp:experiment}
\noindent\textbf{Lighting Consistency (\cref{sec:experiments:synthesis:consistency}). }
We conduct a user study to perceptually evaluate the real-world predictions of our generated images. 
We provide all the samples together with the results in the \path{generation_results.html} of our supplementary material. 

\begin{figure*}[t]
    \centering
    \setlength\tabcolsep{1.25pt}
    \resizebox{0.97\textwidth}{!}{
    \begin{tabular}{cccc|cc}
        \rotatebox{90}{Original Image} &
        \includegraphics[width=0.185\textwidth]{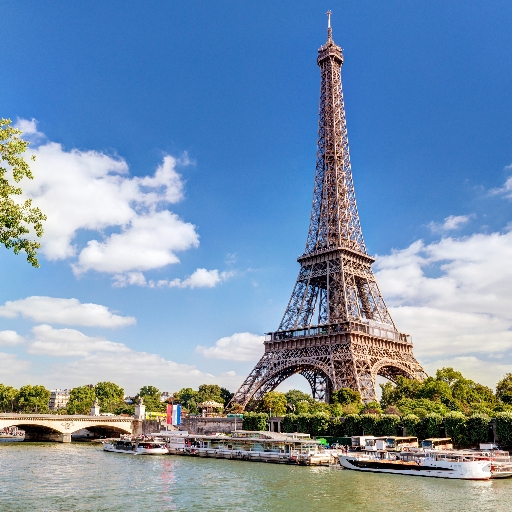} &
        \includegraphics[width=0.185\textwidth]{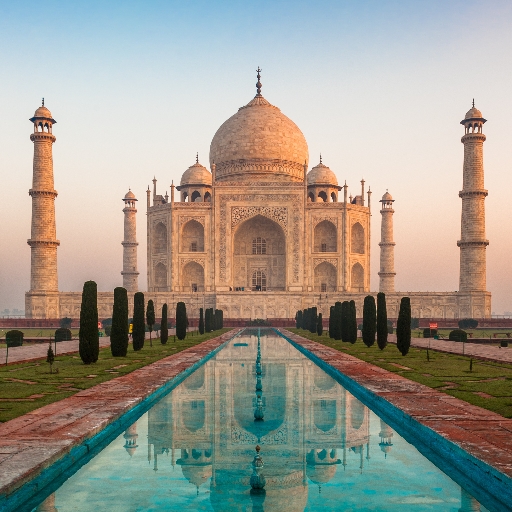} &
        \includegraphics[width=0.185\textwidth]{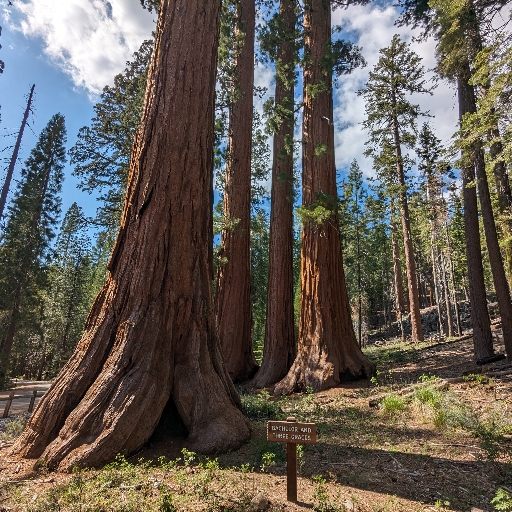} &
        \includegraphics[width=0.185\textwidth]{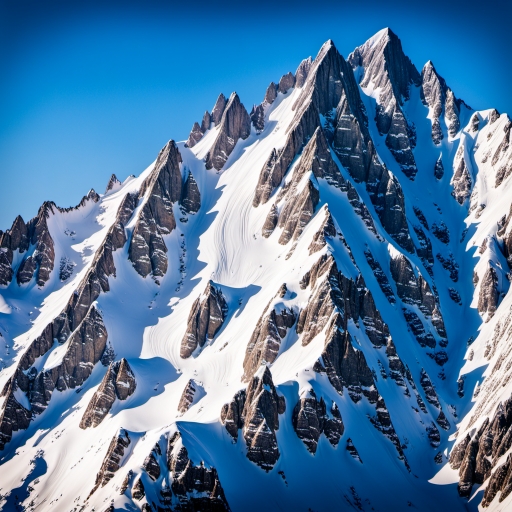} &
        \includegraphics[width=0.185\textwidth]{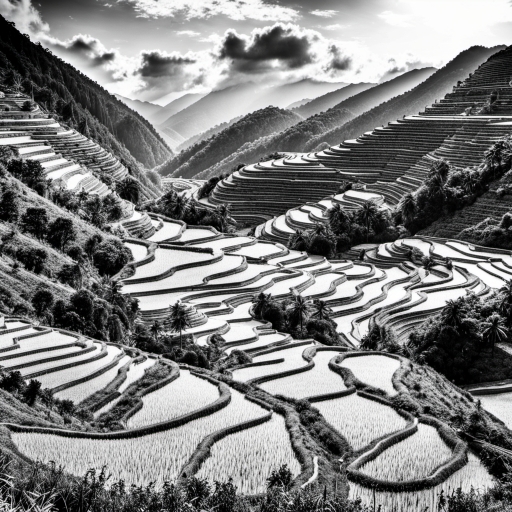} 
        \\

        \rotatebox{90}{Estimated Normals} &
        \includegraphics[width=0.185\textwidth]{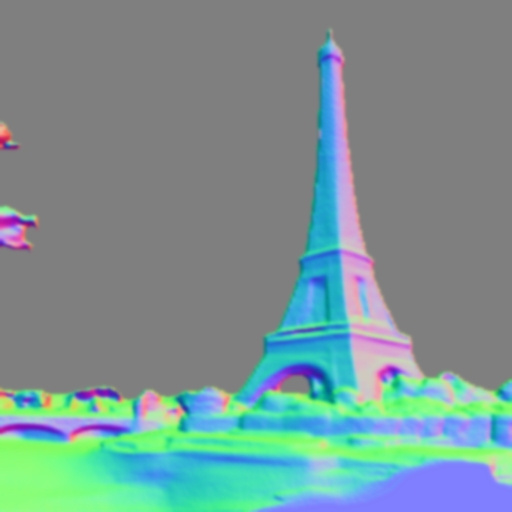} &
        \includegraphics[width=0.185\textwidth]{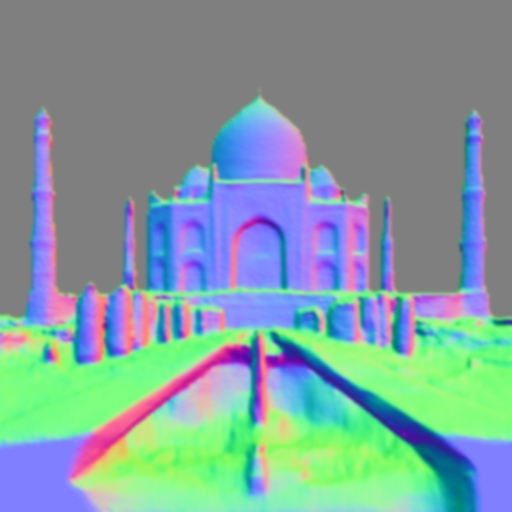} &
        \includegraphics[width=0.185\textwidth]{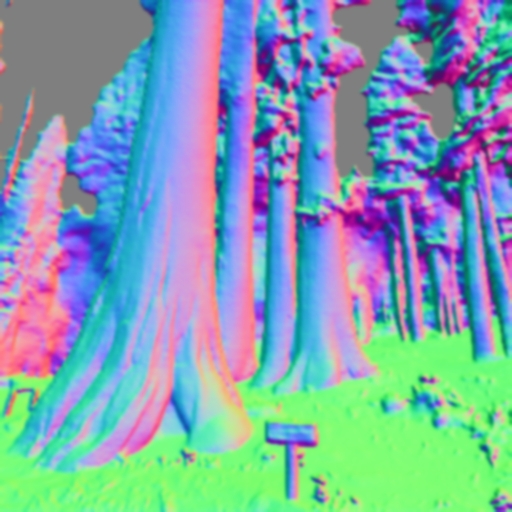} &
        \includegraphics[width=0.185\textwidth]{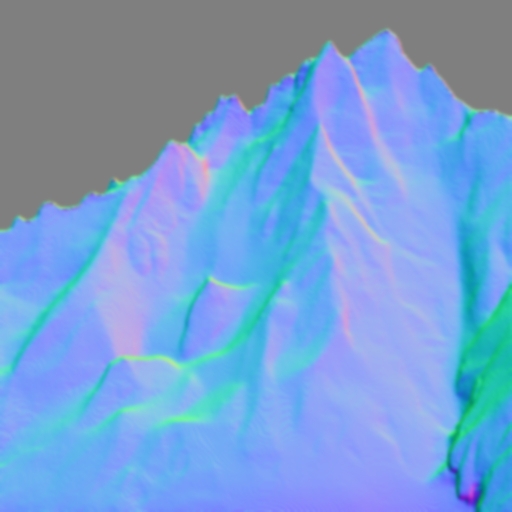} &
        \includegraphics[width=0.185\textwidth]{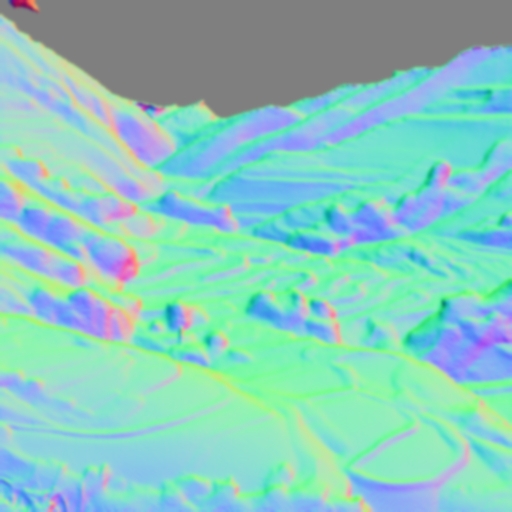} 
        \\

        \raisebox{-12pt}{\rotatebox{90}{Prompt}}&
        \makecell{
            \textit{\footnotesize{Best quality detailed }} \\ 
            \textit{\footnotesize{impressionist painting of}} \\ 
            \textit{\footnotesize{the Eiffel tower and a }} \\ 
            \textit{\footnotesize{blue river in front.}}}  &
        \makecell{
            \textit{\footnotesize{Best quality detailed }} \\ 
            \textit{\footnotesize{medieval painting of}} \\ 
            \textit{\footnotesize{Taj Mahal with fountain.}}}  &
        \makecell{
            \textit{\footnotesize{Best quality detailed}} \\ 
            \textit{\footnotesize{drawing of}} \\
            \textit{\footnotesize{redwood trees.}}}  &
        \makecell{
            \textit{\footnotesize{A snow-covered mountain}} \\ 
            \textit{\footnotesize{peak with evergreen trees}} \\ 
            \textit{\footnotesize{and clear blue sky.}}}  &
        \makecell{
            \textit{\footnotesize{A mountainous landscape with }} \\ 
            \textit{\footnotesize{terraced rice paddies.}}} 
        \\

        \hline
        & \multicolumn{3}{c|}{Real} & \multicolumn{2}{c}{Generated} 
    \end{tabular}
    }
    \vspace{-12pt}
    \caption{\textbf{Out-of-Domain Image Synthesis Details}. 
    We show the original input image together with the estimated normals and text prompts used to generate our images in (\cref{fig:exp:od_controllable_image_synthesis_id}). }
    \label{fig:supp:experiment:od_details}
    \vspace{-15pt}
\end{figure*}

\vspace{6pt}
\noindent\textbf{Lighting Controllability (\cref{sec:experiments:synthesis:controllability}). }
For the samples shown in \Cref{fig:exp:od_controllable_image_synthesis_id}, we used in-the-wild images as well as generated images using SDXL \cite{podell2023sdxl}. 
We show the original images, the estimated normals, and the used text prompts in \cref{fig:supp:experiment:od_details}.
Over the different lighting conditions, we fixed the seed. 
To obtain the SDXL-generated \cite{podell2023sdxl} samples, we used the same prompt as for our generation. 

\vspace{6pt}
\noindent\textbf{Relighting (\cref{sec:experiments:relighting}). }
We conduct a user study to perceptually evaluate the real-world predictions of our relighting application. 
We provide all the samples together with the results in the \path{relighting_results.html} of our supplementary material.

\begin{table}
    \caption{\textbf{Image Cropping Parameters.}
    We crop images from the \cite{hold2019deep} dataset using the following parameters in degrees. 
    }
    \label{tab:cupp:method:dataset:cropping}
    \begin{center}
    \begin{tabular}{l|ccc}
      \toprule
        & Min & Max & Distribution  \\
       \midrule 
      Vertical FOV  & 30 & 110 & Uniform \\
      Azimuth  & 0 & 360 & Uniform \\
      Elevation  & -22.5 & 22.5 & Triangular \\
      Roll  & -22.5 & 22.5 & Triangular \\
      \bottomrule
    \end{tabular}
    \end{center}
\end{table}

\begin{figure*}[t]
    \centering
    \setlength\tabcolsep{0.5pt}
    \fboxsep=0pt
    \resizebox{0.97\textwidth}{!}{
    \begin{tabular}{cccccc}
        \rotatebox{90}{w/ N-dot-L} &
        \begin{tikzpicture}[every node/.style={anchor=north west,inner sep=0pt},x=1pt, y=-1pt,]  
             \node (fig1) at (0,0)
               {\fbox{\includegraphics[width=0.18\textwidth]{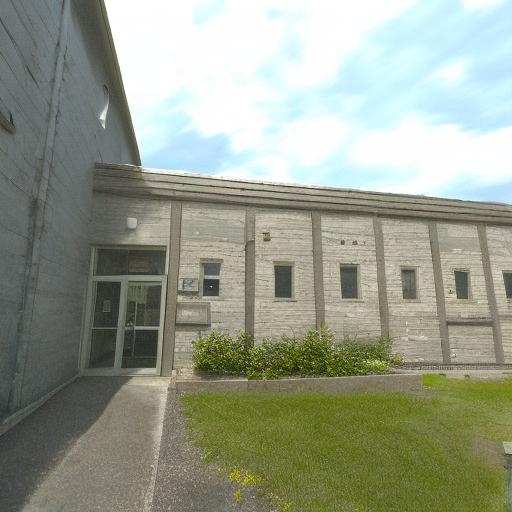}}};   
             \node (fig2) at (-6,-6)
               {\fbox{\includegraphics[width=0.08\textwidth]{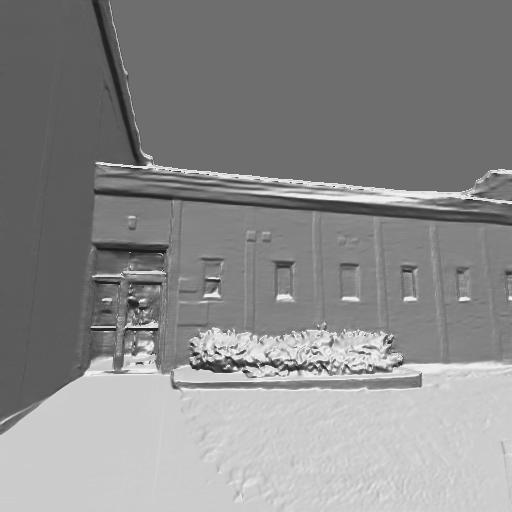}}};
        \end{tikzpicture} &
        \begin{tikzpicture}[every node/.style={anchor=north west,inner sep=0pt},x=1pt, y=-1pt,]  
             \node (fig1) at (0,0)
               {\fbox{\includegraphics[width=0.18\textwidth]{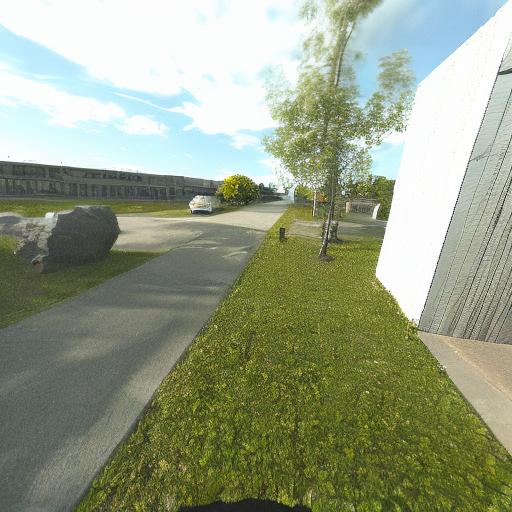}}};   
             \node (fig2) at (-6,-6)
               {\fbox{\includegraphics[width=0.08\textwidth]{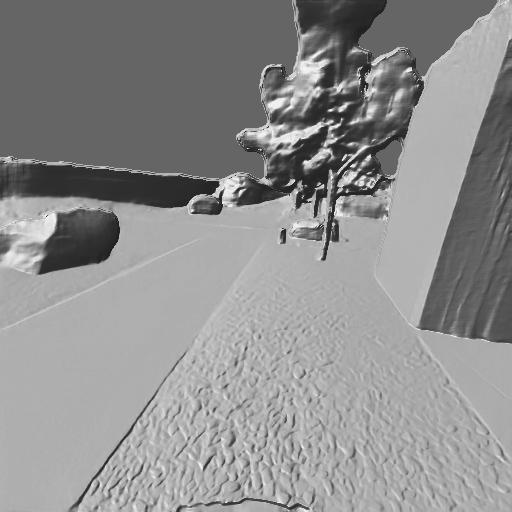}}};
        \end{tikzpicture} &
        \begin{tikzpicture}[every node/.style={anchor=north west,inner sep=0pt},x=1pt, y=-1pt,]  
             \node (fig1) at (0,0)
               {\fbox{\includegraphics[width=0.18\textwidth]{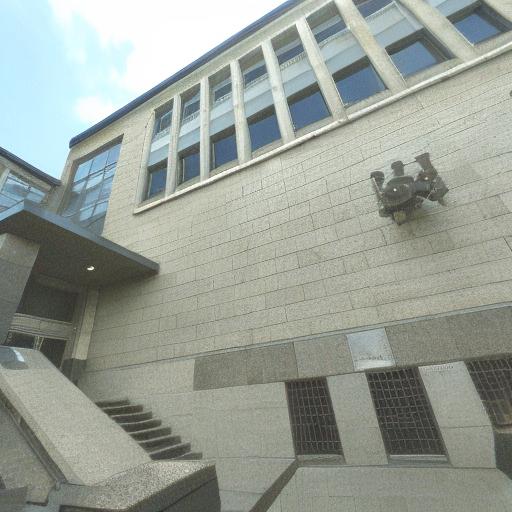}}};   
             \node (fig2) at (-6,-6)
               {\fbox{\includegraphics[width=0.08\textwidth]{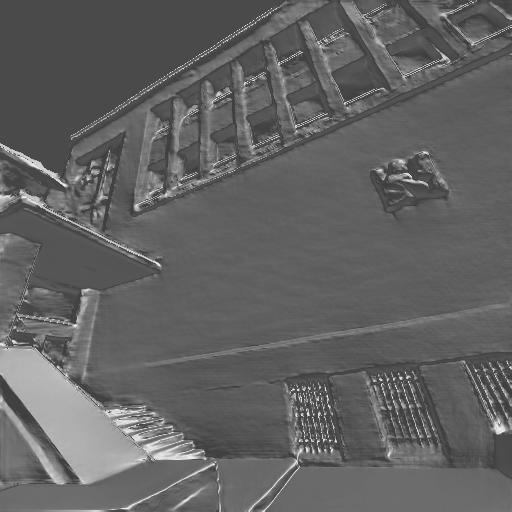}}};
        \end{tikzpicture} &
        \begin{tikzpicture}[every node/.style={anchor=north west,inner sep=0pt},x=1pt, y=-1pt,]  
             \node (fig1) at (0,0)
               {\fbox{\includegraphics[width=0.18\textwidth]{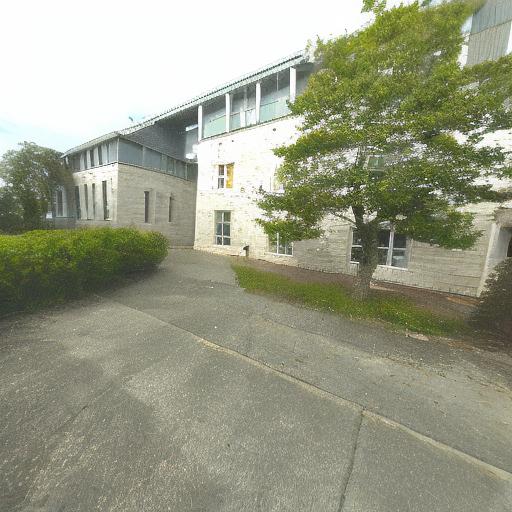}}};   
             \node (fig2) at (-6,-6)
               {\fbox{\includegraphics[width=0.08\textwidth]{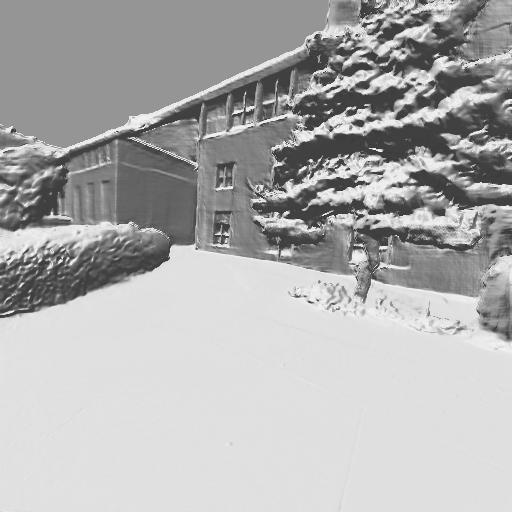}}};
        \end{tikzpicture} &
        \begin{tikzpicture}[every node/.style={anchor=north west,inner sep=0pt},x=1pt, y=-1pt,]  
             \node (fig1) at (0,0)
               {\fbox{\includegraphics[width=0.18\textwidth]{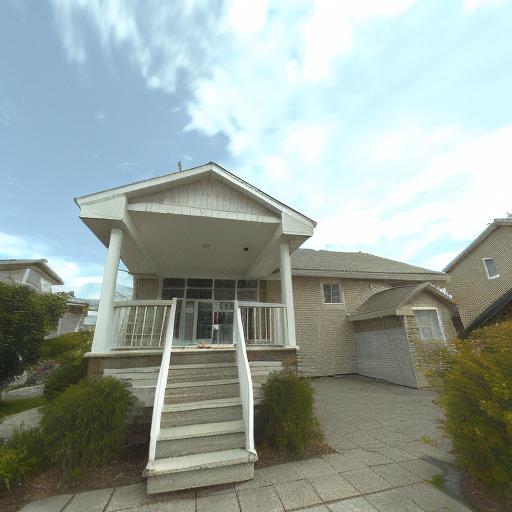}}};   
             \node (fig2) at (-6,-6)
               {\fbox{\includegraphics[width=0.08\textwidth]{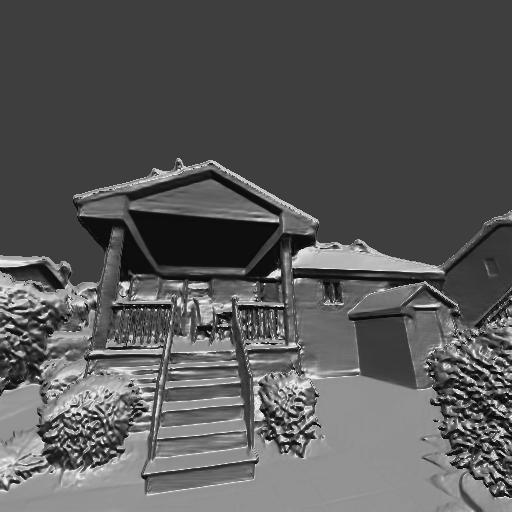}}};
        \end{tikzpicture} 
        \\

        \rotatebox{90}{Ours} &
        \begin{tikzpicture}[every node/.style={anchor=north west,inner sep=0pt},x=1pt, y=-1pt,]  
             \node (fig1) at (0,0)
               {\fbox{\includegraphics[width=0.18\textwidth]{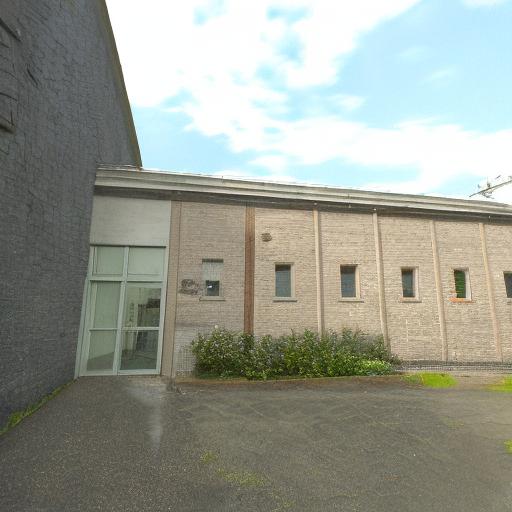}}};   
             \node (fig2) at (-6,-6)
               {\fbox{\includegraphics[width=0.08\textwidth]{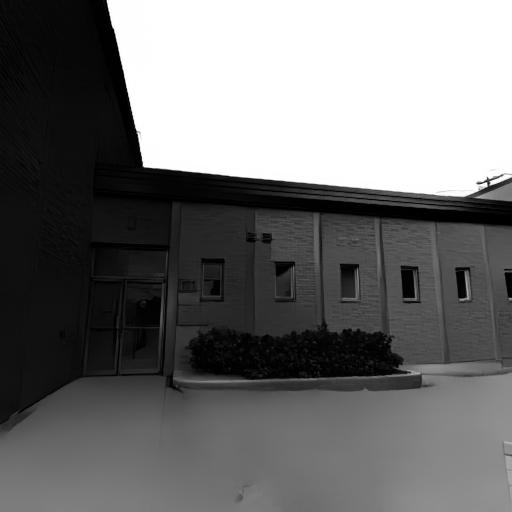}}};
        \end{tikzpicture} &
        \begin{tikzpicture}[every node/.style={anchor=north west,inner sep=0pt},x=1pt, y=-1pt,]  
             \node (fig1) at (0,0)
               {\fbox{\includegraphics[width=0.18\textwidth]{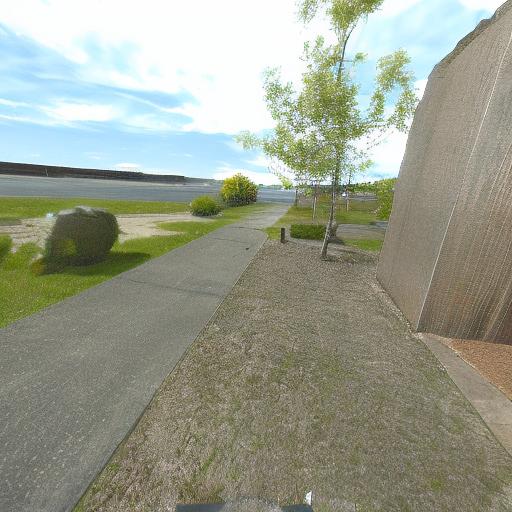}}};   
             \node (fig2) at (-6,-6)
               {\fbox{\includegraphics[width=0.08\textwidth]{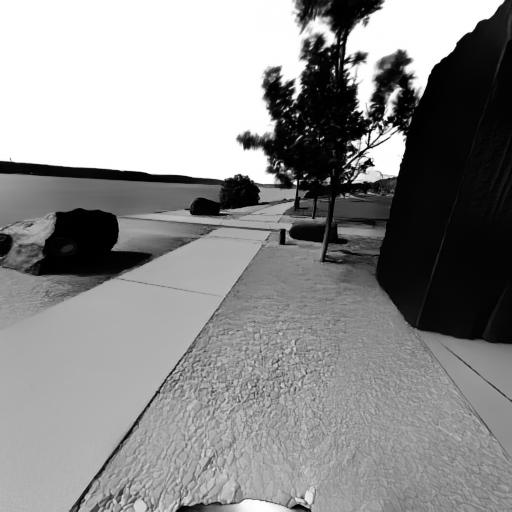}}};
        \end{tikzpicture} &
        \begin{tikzpicture}[every node/.style={anchor=north west,inner sep=0pt},x=1pt, y=-1pt,]  
             \node (fig1) at (0,0)
               {\fbox{\includegraphics[width=0.18\textwidth]{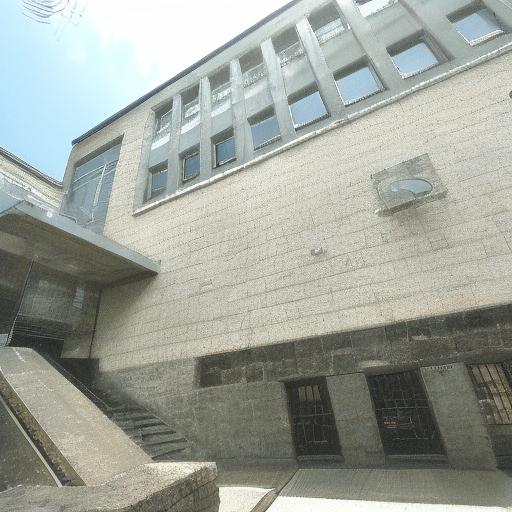}}};   
             \node (fig2) at (-6,-6)
               {\fbox{\includegraphics[width=0.08\textwidth]{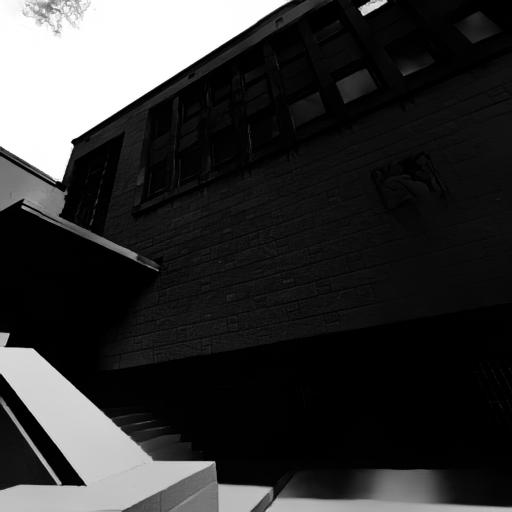}}};
        \end{tikzpicture}  &
        \begin{tikzpicture}[every node/.style={anchor=north west,inner sep=0pt},x=1pt, y=-1pt,]  
             \node (fig1) at (0,0)
               {\fbox{\includegraphics[width=0.18\textwidth]{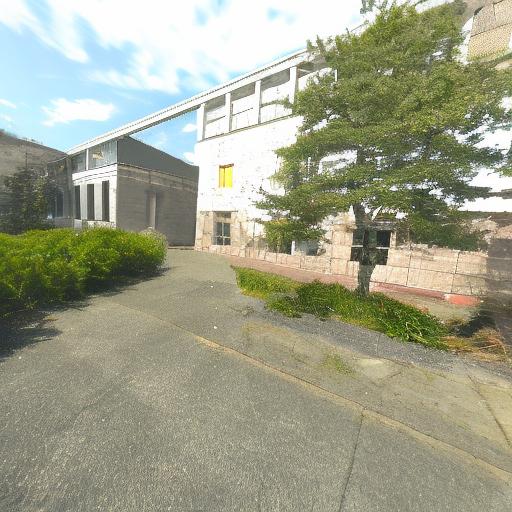}}};   
             \node (fig2) at (-6,-6)
               {\fbox{\includegraphics[width=0.08\textwidth]{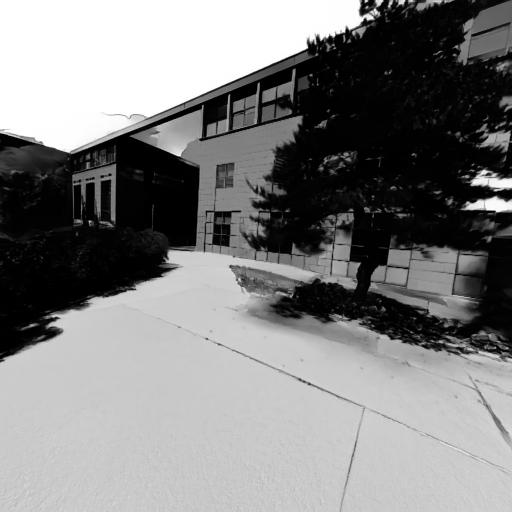}}};
        \end{tikzpicture}   &
        \begin{tikzpicture}[every node/.style={anchor=north west,inner sep=0pt},x=1pt, y=-1pt,]  
             \node (fig1) at (0,0)
               {\fbox{\includegraphics[width=0.18\textwidth]{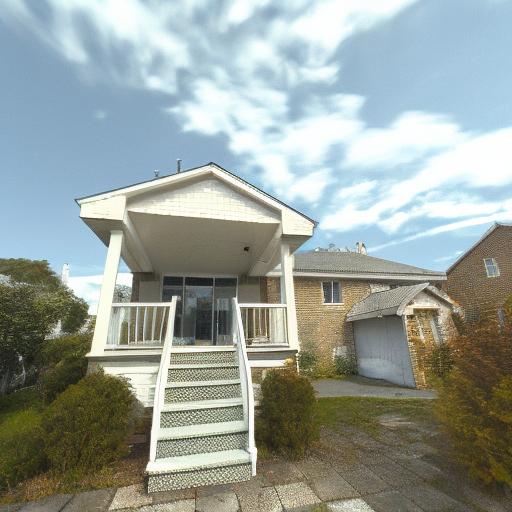}}};   
             \node (fig2) at (-6,-6)
               {\fbox{\includegraphics[width=0.08\textwidth]{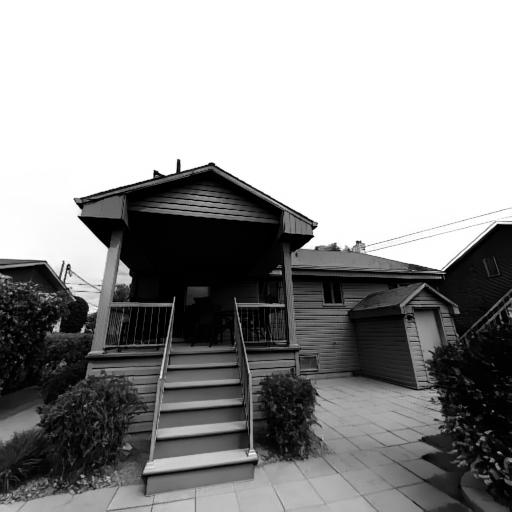}}};
        \end{tikzpicture} 

    \end{tabular}
    } 
    
    \vspace{-12pt}
    \caption{\textbf{Effect of Lighting Representation}. 
    We show additional results to our ablation about the effect of cast shadows (\cref{fig:exp:abl_lighting_representation}). 
    \readyForFeedback}
    \label{fig:supp:exp:results:abl_lighting_representation}
    \vspace{-15pt}
\end{figure*}

\section{Additional Results}
\label{sec:supp:results}
\noindent\textbf{Effect of Lighting Representation (\cref{sec:experiments:synthesis:ablation}). }
Our method uses direct shading, which contains information about cast shadows.
Since SD \cite{rombach2022high} does not model light transport, cast shadows provide valuable information to the model. 
We show additional samples to our ablation (\cref{fig:exp:abl_lighting_representation}) in \cref{fig:supp:exp:results:abl_lighting_representation}.

\begin{figure*}[t]
    \centering
    \setlength\tabcolsep{1pt}
    \fboxsep=0pt
    \resizebox{0.97\textwidth}{!}{
    \begin{tabular}{c|cccc}
        \begin{tikzpicture}[every node/.style={anchor=north west,inner sep=0pt},x=0.5pt, y=-0.5pt,]  
             \node (fig1) at (0,0)
               {\fbox{\adjincludegraphics[trim={0 {.3\width} 0 0},clip,width=0.17\linewidth]{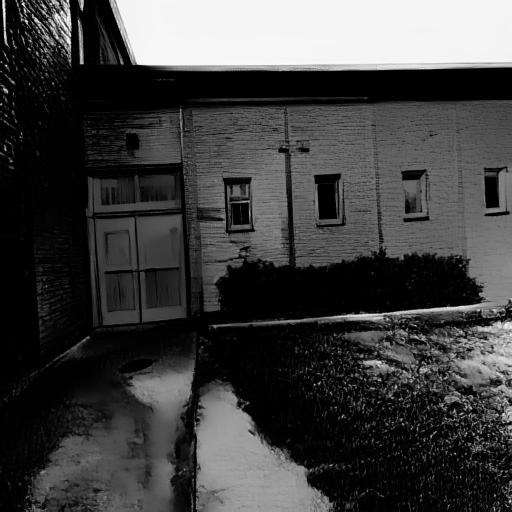}}};   
             \node (fig2) at (-6,-6)
               {\fbox{\adjincludegraphics[trim={0 {.3\width} 0 0},clip,width=0.08\linewidth]{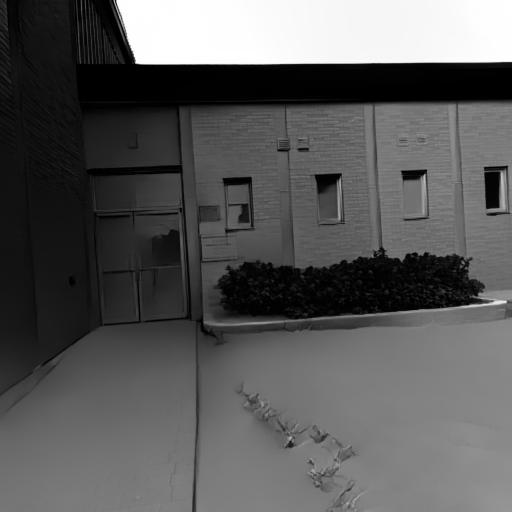}}};
        \end{tikzpicture} &
        \begin{tikzpicture}[every node/.style={anchor=north west,inner sep=0pt},x=0.5pt, y=-0.5pt,]  
             \node (fig1) at (0,0)
               {\fbox{\adjincludegraphics[trim={0 {.3\width} 0 0},clip,width=0.17\linewidth]{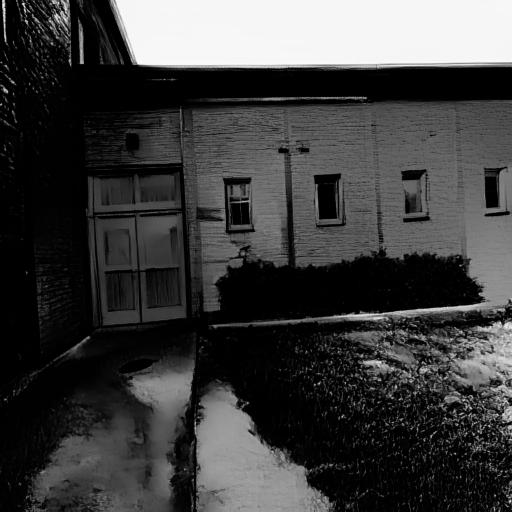}}};   
             \node (fig2) at (-6,-6)
               {\fbox{\adjincludegraphics[trim={0 {.3\width} 0 0},clip,width=0.08\linewidth]{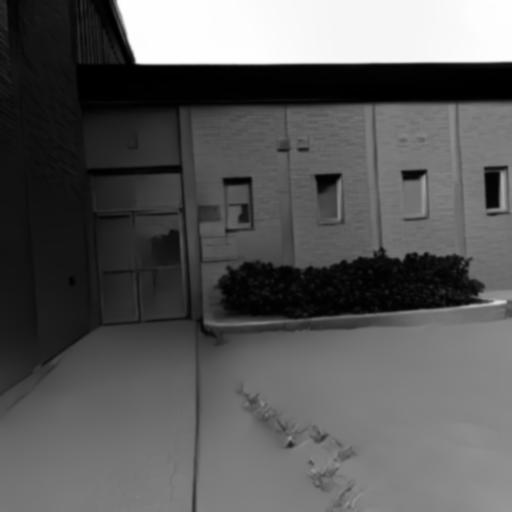}}};
        \end{tikzpicture} &
        \begin{tikzpicture}[every node/.style={anchor=north west,inner sep=0pt},x=0.5pt, y=-0.5pt,]  
             \node (fig1) at (0,0)
               {\fbox{\adjincludegraphics[trim={0 {.3\width} 0 0},clip,width=0.17\linewidth]{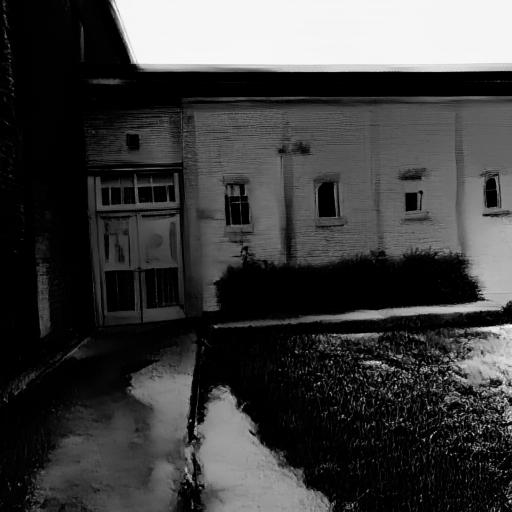}}};   
             \node (fig2) at (-6,-6)
               {\fbox{\adjincludegraphics[trim={0 {.3\width} 0 0},clip,width=0.08\linewidth]{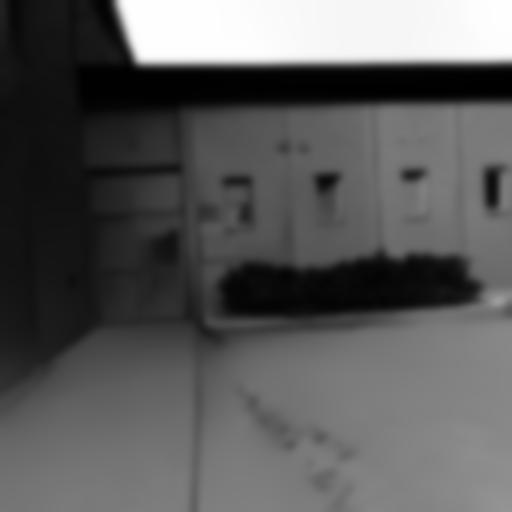}}};
        \end{tikzpicture} &
        \begin{tikzpicture}[every node/.style={anchor=north west,inner sep=0pt},x=0.5pt, y=-0.5pt,]  
             \node (fig1) at (0,0)
               {\fbox{\adjincludegraphics[trim={0 {.3\width} 0 0},clip,width=0.17\linewidth]{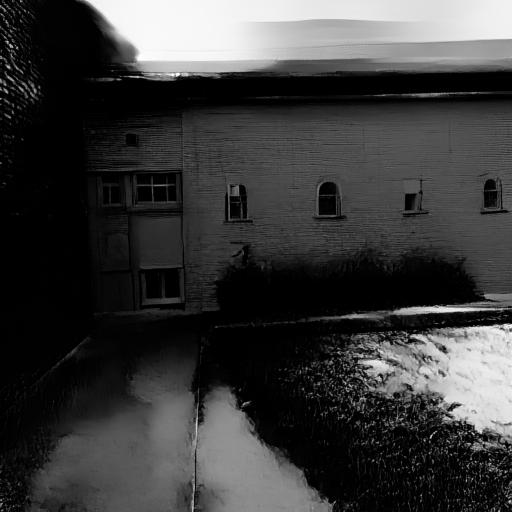}}};   
             \node (fig2) at (-6,-6)
               {\fbox{\adjincludegraphics[trim={0 {.3\width} 0 0},clip,width=0.08\linewidth]{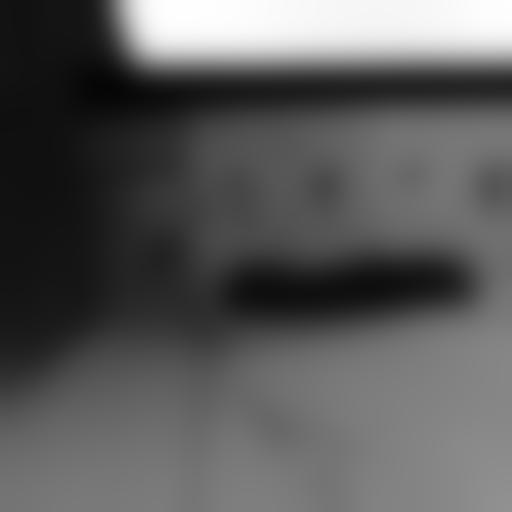}}};
        \end{tikzpicture} &
        \begin{tikzpicture}[every node/.style={anchor=north west,inner sep=0pt},x=0.5pt, y=-0.5pt,]  
             \node (fig1) at (0,0)
               {\fbox{\adjincludegraphics[trim={0 {.3\width} 0 0},clip,width=0.17\linewidth]{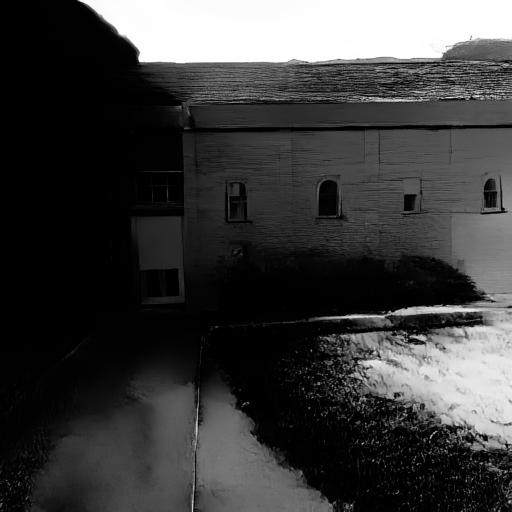}}};   
             \node (fig2) at (-6,-6)
               {\fbox{\adjincludegraphics[trim={0 {.3\width} 0 0},clip,width=0.08\linewidth]{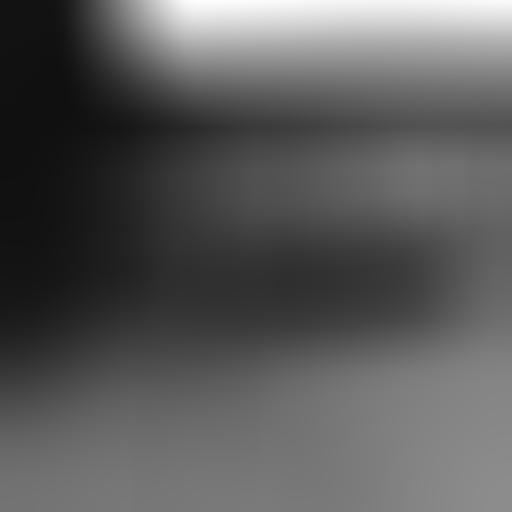}}};
        \end{tikzpicture} 
        \\[-6pt]

        \scriptsize{Ours: \textbf{12.69}dB} &
        \scriptsize{$\sigma$}1: \textbf{12.58}dB  &
        \scriptsize{$\sigma$}4: \textbf{11.52}dB &
        \scriptsize{$\sigma$}16: \textbf{10.48}dB &
        \scriptsize{$\sigma$}64: \textbf{10.14}dB \\

        \begin{tikzpicture}[every node/.style={anchor=north west,inner sep=0pt},x=0.5pt, y=-0.5pt,]  
             \node (fig1) at (0,0)
               {\fbox{\adjincludegraphics[trim={0 {.3\width} 0 0},clip,width=0.17\linewidth]{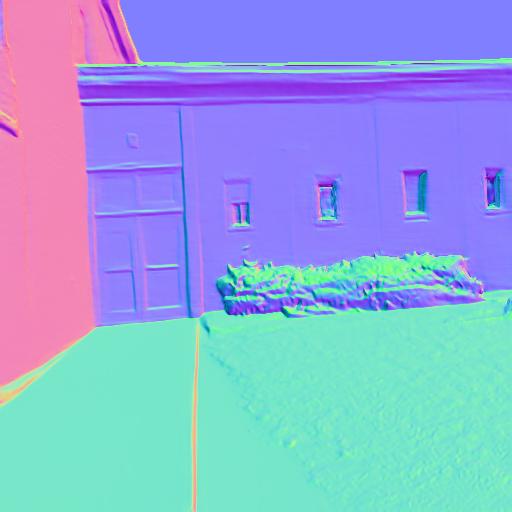}}};   
             \node (fig2) at (-6,-6)
               {\fbox{\adjincludegraphics[trim={0 {.3\width} 0 0},clip,width=0.08\linewidth]{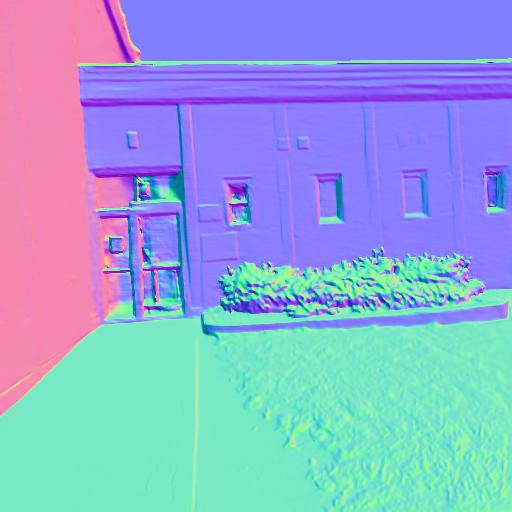}}};
        \end{tikzpicture} &
        \begin{tikzpicture}[every node/.style={anchor=north west,inner sep=0pt},x=0.5pt, y=-0.5pt,]  
             \node (fig1) at (0,0)
               {\fbox{\adjincludegraphics[trim={0 {.3\width} 0 0},clip,width=0.17\linewidth]{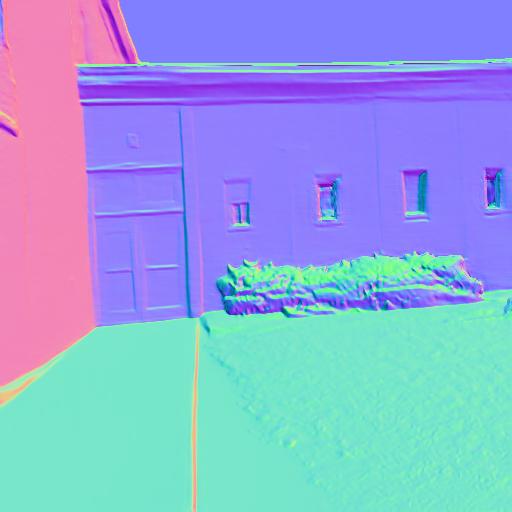}}};   
             \node (fig2) at (-6,-6)
               {\fbox{\adjincludegraphics[trim={0 {.3\width} 0 0},clip,width=0.08\linewidth]{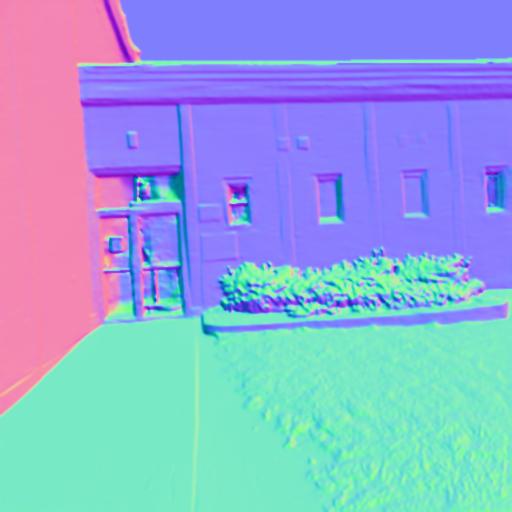}}};
        \end{tikzpicture} &
        \begin{tikzpicture}[every node/.style={anchor=north west,inner sep=0pt},x=0.5pt, y=-0.5pt,]  
             \node (fig1) at (0,0)
               {\fbox{\adjincludegraphics[trim={0 {.3\width} 0 0},clip,width=0.17\linewidth]{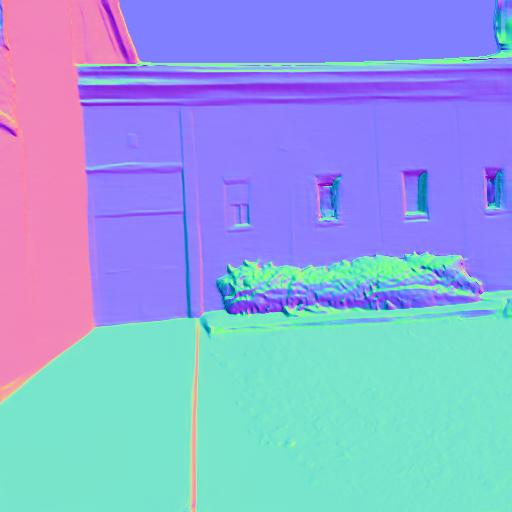}}};   
             \node (fig2) at (-6,-6)
               {\fbox{\adjincludegraphics[trim={0 {.3\width} 0 0},clip,width=0.08\linewidth]{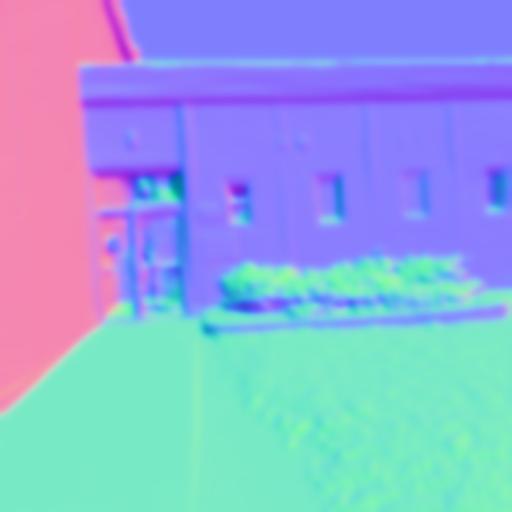}}};
        \end{tikzpicture} &
        \begin{tikzpicture}[every node/.style={anchor=north west,inner sep=0pt},x=0.5pt, y=-0.5pt,]  
             \node (fig1) at (0,0)
               {\fbox{\adjincludegraphics[trim={0 {.3\width} 0 0},clip,width=0.17\linewidth]{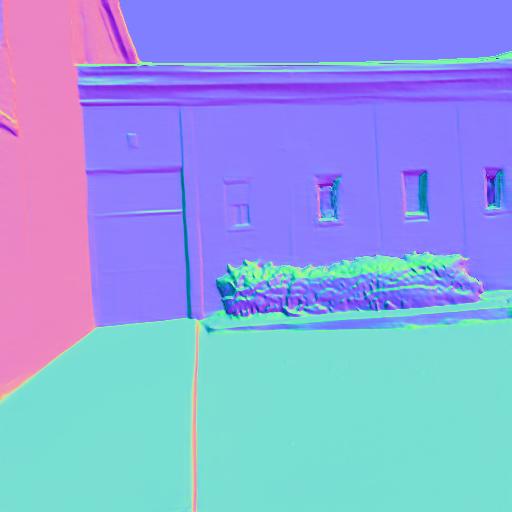}}};   
             \node (fig2) at (-6,-6)
               {\fbox{\adjincludegraphics[trim={0 {.3\width} 0 0},clip,width=0.08\linewidth]{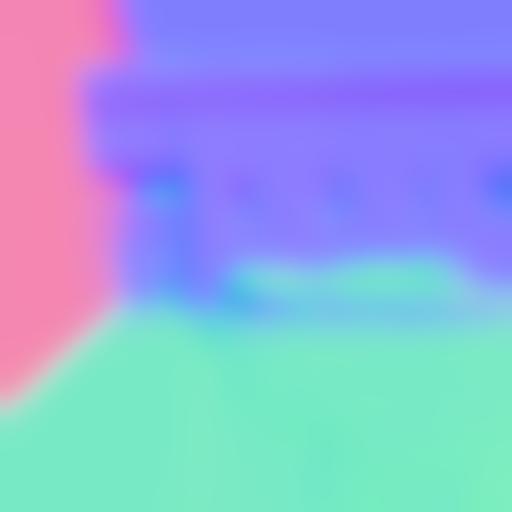}}};
        \end{tikzpicture} &
        \begin{tikzpicture}[every node/.style={anchor=north west,inner sep=0pt},x=0.5pt, y=-0.5pt,]  
             \node (fig1) at (0,0)
               {\fbox{\adjincludegraphics[trim={0 {.3\width} 0 0},clip,width=0.17\linewidth]{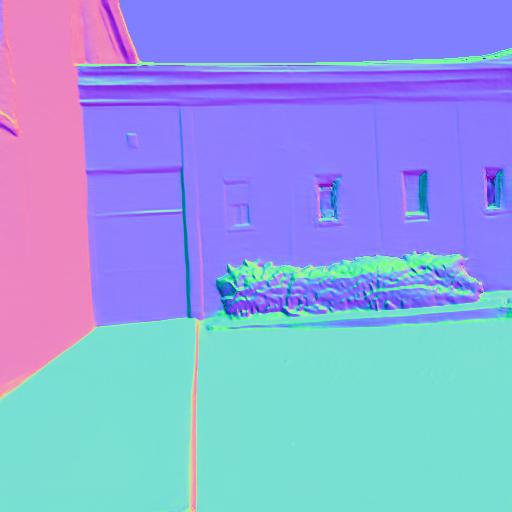}}};   
             \node (fig2) at (-6,-6)
               {\fbox{\adjincludegraphics[trim={0 {.3\width} 0 0},clip,width=0.08\linewidth]{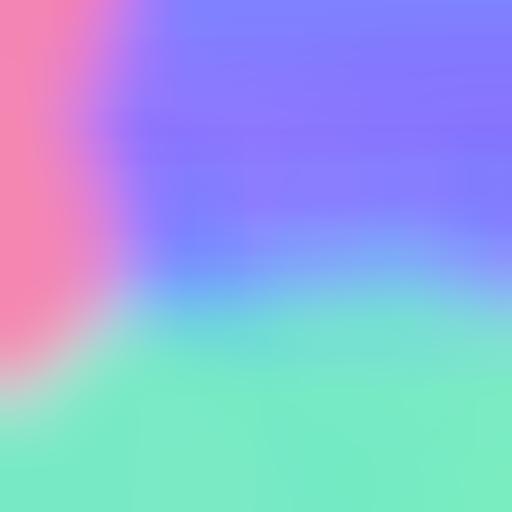}}};
        \end{tikzpicture} 
        \\[-6pt]

        \scriptsize{Ours: \textbf{17.47}dB} &
        \scriptsize{$\sigma$}1: \textbf{17.47}dB  &
        \scriptsize{$\sigma$}4: \textbf{17.41}dB &
        \scriptsize{$\sigma$}16: \textbf{17.21}dB &
        \scriptsize{$\sigma$}64: \textbf{17.20}dB \\ 
    \end{tabular}
    }
    \vspace{-12pt}
    \caption{\textbf{Control robustness}. 
    Test set average for original control reconstruction of our generated samples with increasingly blurred input (top left insets). 
    }
    \label{fig:supp:robustness}
    \vspace{-15pt}
\end{figure*}

\noindent\textbf{Control robustness (\cref{sec:method:dataset}). }
\revise{Our method is trained on automatically generated normal and shading maps.
In \cref{fig:supp:robustness}, we evaluate the robustness of our method against noisy control signals. 
We perturb the control signal by blurring it and evaluate the consistency, compared to the original control signal, as done in \cref{tab:exp:control_consistency}.
Even with highly perturbed input, our method yield consistent results by successfully utilizing information from the other control.
}

\end{document}